\documentclass{article} 
\usepackage{iclr2026_conference,times}

\usepackage{amsmath,amsfonts,bm}

\def\eqref#1{equation~\ref{#1}}

\def\1{\bm{1}}

\DeclareMathAlphabet{\mathsfit}{\encodingdefault}{\sfdefault}{m}{sl}
\SetMathAlphabet{\mathsfit}{bold}{\encodingdefault}{\sfdefault}{bx}{n}

\usepackage{float}
\usepackage{wrapfig}   
\usepackage[hidelinks]{hyperref}
\usepackage{url}
\usepackage{enumitem}
\usepackage[most]{tcolorbox}
\usepackage{fontawesome} 
\usepackage{booktabs}
\usepackage[table]{xcolor}
\usepackage{tabularx}
\usepackage{array}
\usepackage{xurl}
\usepackage{amsmath}
\usepackage{amssymb}
\usepackage{mathtools}
\usepackage{amsthm}
\usepackage{listings}
\usepackage{subcaption}
\usepackage{needspace}
\definecolor{color11}{rgb}{1, 0.8, 0.8}  
\definecolor{color12}{rgb}{1, 0.9, 0.9}  
\definecolor{color21}{RGB}{255, 224, 127}  
\definecolor{color22}{RGB}{255, 239, 179}  
\definecolor{color31}{RGB}{198, 230, 195}  
\definecolor{color32}{RGB}{224, 239, 225}  

\newcommand{\tablestyle}[2]{\setlength{\tabcolsep}{#1}\renewcommand{\arraystretch}{#2}\centering\footnotesize}

\newcounter{prompt}
\newenvironment{promptbox}[2][]{%
  \refstepcounter{prompt}%
  \begin{tcolorbox}[
    enhanced,
    breakable,
    colback=white,
    colframe=purple!70!blue!80!black,
    coltitle=white,
    title={Prompt~\theprompt: #2},
    halign title=flush left,
    fonttitle=\bfseries,
    fontupper=\footnotesize,
    boxrule=0.7pt,
    arc=1pt,
    left=4pt,
    right=4pt,
    top=3pt,
    bottom=3pt,
    before skip=0.45em,
    after skip=0.45em,
    #1]
  \setlength{\parindent}{0pt}%
  \setlength{\parskip}{0.25em}%
  \setlength{\leftmargini}{1.2em}%
  \setlength{\leftmarginii}{1.2em}%
}{\end{tcolorbox}}
\newenvironment{promptbox2}[2][]{%
  \refstepcounter{prompt}%
  \begin{tcolorbox}[
    enhanced,
    breakable,
    colback=white,
    colframe=blue!50!black,
    coltitle=white,
    title={Prompt~\theprompt: #2},
    halign title=flush left,
    fonttitle=\bfseries,
    fontupper=\footnotesize,
    boxrule=0.7pt,
    arc=1pt,
    left=4pt,
    right=4pt,
    top=3pt,
    bottom=3pt,
    before skip=0.45em,
    after skip=0.45em,
    #1]
  \setlength{\parindent}{0pt}%
  \setlength{\parskip}{0.25em}%
  \setlength{\leftmargini}{1.2em}%
  \setlength{\leftmarginii}{1.2em}%
}{\end{tcolorbox}}

\title{CS-54k: A Paper-Grounded Corpus and Benchmark for AI Research QA}

\author{%
\begin{tabular}{@{}c@{}}
\textbf{Penghao Wang}$^{1}$\enspace
\textbf{Yuhao Zhou}$^{1}$\footnotemark[4]\enspace 
\textbf{Mengxuan Wu}$^{1}$\enspace 
\textbf{Ziheng Qin}$^{1}$\enspace 
\textbf{Bangyuan Zhu}$^{2}$\\
\textbf{Shengbin Huang}$^{3}$\enspace
\textbf{Xuanlei Zhao}$^{1}$\enspace
\textbf{Panpan Zhang}$^{1}$\enspace
\textbf{Xiaojiang Peng}$^{3}$\enspace
\textbf{Yuzhang Shang}$^{4}$\\
\textbf{Jianfei Yang}$^{3}$\enspace
\textbf{Zheng Zhu}$^{5}$\enspace
\textbf{Tianlong Chen}$^{6}$\enspace
\textbf{Zhangyang Wang}$^{7}$\enspace
\textbf{Kai Wang}$^{1}$\footnotemark[2]\\[4pt]
$^{1}$NUS \enspace
$^{2}$NTU \enspace
$^{3}$SZTU \enspace
$^{4}$UCF \enspace 
$^{5}$GigaAI \enspace
$^{6}$UNC \enspace
$^{7}$UT Austin 
\end{tabular}%
}

\iclrfinalcopy 
\begin{document}

\maketitle
\fancyhead{}
\renewcommand{\headrulewidth}{0pt}
\renewcommand{\thefootnote}{\fnsymbol{footnote}}
\footnotetext[4]{Project lead.} \footnotetext[2]{Corresponding author.}
\begin{abstract}
Large language models (LLMs) are increasingly being used to support computer science research.
To contribute reliably, these systems must be able to work with information distributed across research papers and across different aspects of the research process.
However, large-scale resources that systematically organize such information into structured supervision for both model evaluation and post-training remain limited.
To address this gap, we propose a systematic paper-to-QA construction pipeline that organizes information from AI research papers into structured QA pairs supported by source evidence.
Using this pipeline, we construct \textbf{CS-54k}, a 54k-pair corpus derived from
14,474 openly licensed AI research papers.
The corpus covers eight recurring paper components, including research domain, prior work, existing challenges, motivation, findings and assumptions, methods, experimental settings, and experimental results.
We further derive two complementary subsets:
\textbf{CS-4k}, a benchmark for evaluating LLMs on paper-derived research QA, and
\textbf{CS-50k}, a companion corpus for post-training.
Extensive evaluations on CS-4k reveal substantial performance differences across existing LLMs and paper components, with experimental settings and results remaining particularly challenging.
Moreover, post-training on CS-50k substantially improves \texttt{Qwen2.5-7B-Instruct} on CS-4k, demonstrating the practical utility of structured, paper-derived supervision.
Together, CS-4k and CS-50k provide a unified and scalable resource for benchmarking and improving LLMs on research paper question answering.
Our code and dataset are available at \mbox{\url{https://github.com/wph6/ResearchGPT}} and \mbox{\url{https://huggingface.co/datasets/wph6/CS-54k}}, respectively.
\end{abstract}

\section{Introduction}
\label{sec:intro}

Large language models (LLMs) are increasingly participating in computer science research, supporting
activities ranging from literature analysis to experimental workflows
\citep{luo2025llm4sr,zhang2025exploring,eger2025transforming,lu2026towards}.
However, research papers largely remain external sources of information: document access alone does
not provide a systematic way to turn their contents into model capabilities that can be consistently
evaluated and improved.
To bridge this gap, we study a data-centric approach that transforms information distributed across
research papers into structured QA supervision, making it explicitly queryable, verifiable against
source evidence, and directly usable for model evaluation and post-training.

Existing benchmarks for LLMs in scientific research remain fragmented and are often misaligned with
how information is organized in research papers.
Broadly, prior efforts fall into two categories: \emph{evaluation-as-exam} and
\emph{evaluation-as-agent}.
Exam-style benchmarks, such as SuperGPQA \citep{du2025supergpqa} and HLE \citep{phan2025HLE},
primarily assess broad factual knowledge and general reasoning.
Agent-style benchmarks, such as CSR-Bench \citep{xiao2025csr}, DataSciBench
\citep{zhang2025datascibench}, and DSBench \citep{jing2024dsbench}, instead focus on workflow
execution, including data analysis, code deployment, and repository-level tasks.
These resources are valuable, but they are not designed as large-scale corpora of paper-derived
questions that systematically span paper components while jointly supporting
model evaluation and post-training.
This gap motivates a unified benchmark and training corpus for research paper question answering.

\begin{figure}[t]
    \centering
    \begin{minipage}[b]{0.36\textwidth}
        \centering
        \includegraphics[width=\linewidth]{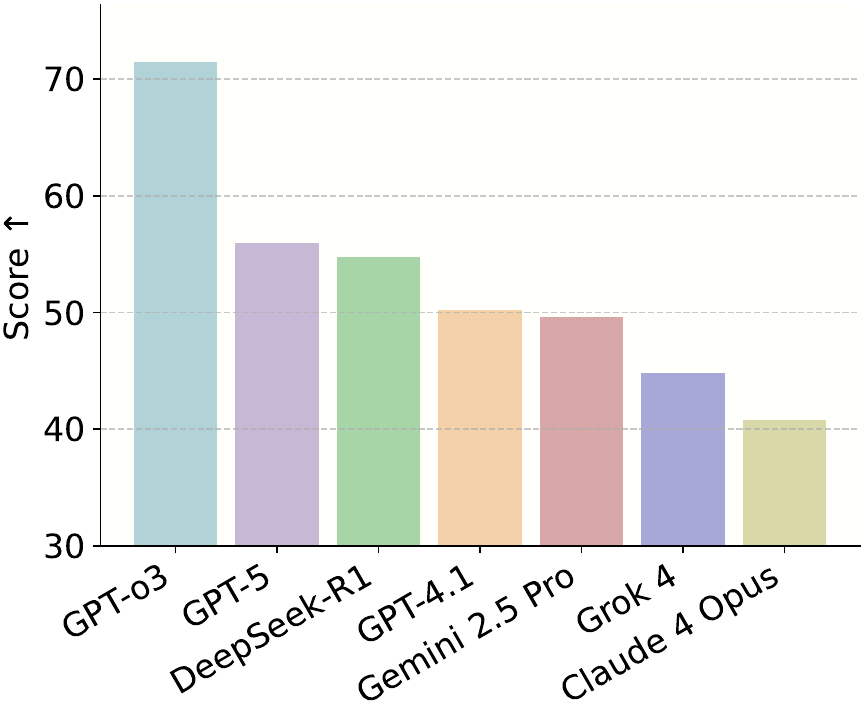}
    \end{minipage}
    \hfill
    \begin{minipage}[b]{0.58\textwidth}
        \centering
        \includegraphics[width=\linewidth]{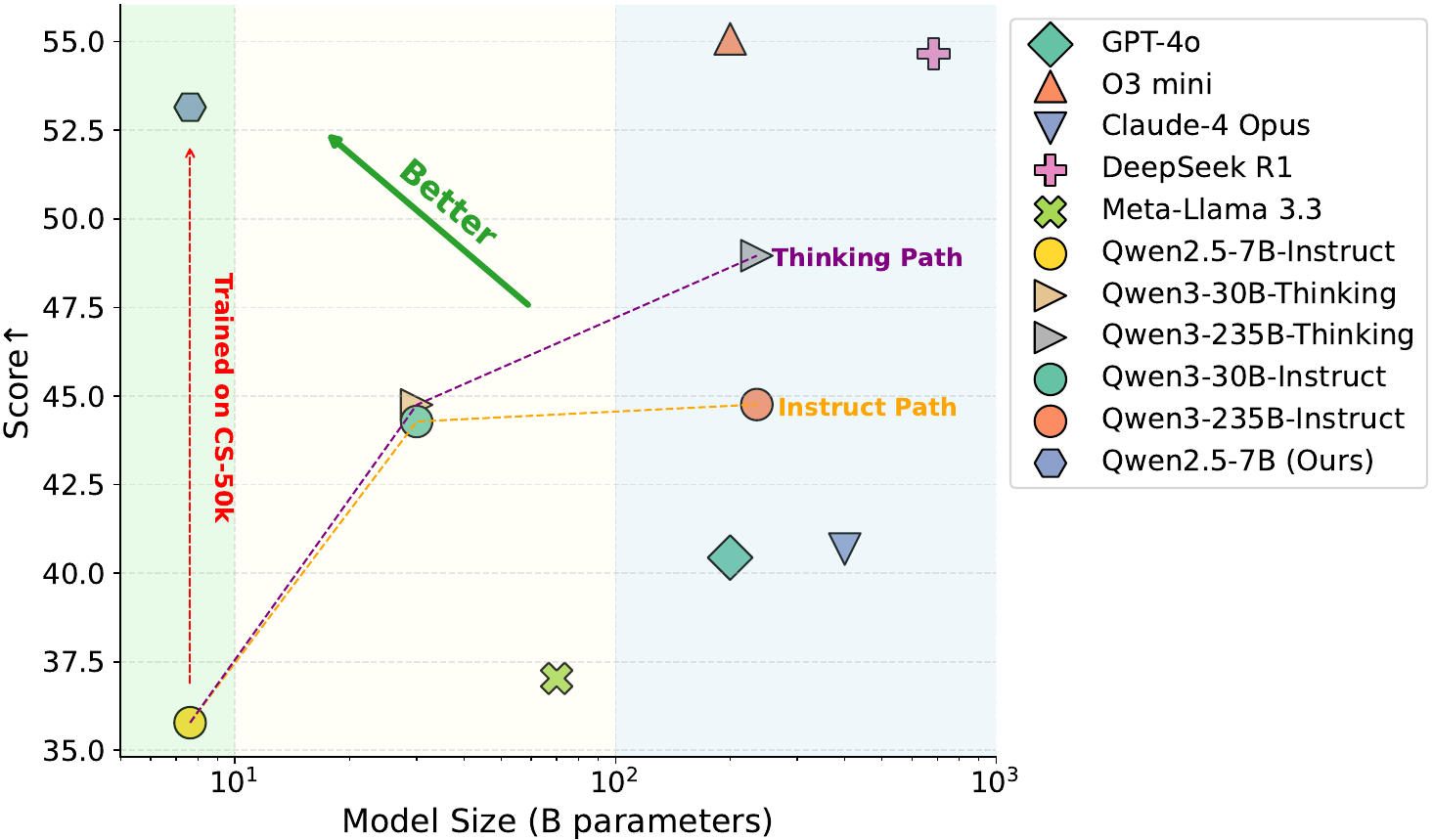}
    \end{minipage}
    \caption{\textbf{Overview of CS-54k evaluation and post-training studies.}
    \textbf{Left:} performance of selected existing LLMs on CS-4k; these models are evaluated
    without being trained on CS-50k in our experiments.
    \textbf{Right:} model size vs.\ score, with the trained Qwen result illustrating the utility of
    CS-50k for post-training.}
    \label{fig:cs4k_results}
\end{figure}

To address this gap, we construct \textbf{CS-54k}, a corpus of approximately 54,000 QA pairs derived
from 14,474 openly licensed AI research papers.
Each question is derived from a specific paper and paired with a reference answer supported by
evidence from that paper.
The corpus is organized around eight recurring paper components: research domain, previous
methods, existing challenges, motivation, findings or assumptions, methods, experimental settings,
and experimental results.
Organizing the corpus by paper component turns information distributed throughout research papers
into a standardized interface through which model performance can be evaluated, diagnosed, and
improved.

We construct CS-54k through a reproducible paper-to-QA pipeline combining document parsing,
hierarchical chunking, retrieval-grounded question and answer generation, automatic filtering, and
expert validation.
Unlike generic QA generation, the pipeline organizes paper information according to recurring
paper components before generating questions and answers.
This design provides systematic coverage of different information types while keeping the generated
answers grounded in source evidence.

From the unified corpus, we derive two complementary subsets.
\textbf{CS-4k} serves as a benchmark for characterizing existing LLMs on paper-derived research QA,
including differences in performance across components.
\textbf{CS-50k} serves as a companion training corpus for examining whether structured,
paper-derived supervision improves \texttt{Qwen2.5-\allowbreak 7B-\allowbreak Instruct}
\citep{qwen2,qwen2.5}.
CS-4k and CS-50k contain disjoint QA instances drawn from the same paper collection; accordingly, the
post-training study evaluates the practical utility of CS-50k as supervision within this
paper-derived QA setting.

Extensive evaluations show that CS-4k reveals substantial performance variation across existing LLMs
and paper components, with experimental settings and results remaining particularly
challenging.
Post-training on CS-50k substantially improves \texttt{Qwen2.5-7B-Instruct} on CS-4k, demonstrating
that the resulting corpus provides effective supervision for research paper QA.

Our main contributions are summarized as follows:

\begin{itemize}

    \item We introduce \textbf{CS-54k}, a unified paper-derived QA resource consisting of
    \textbf{CS-4k}, a benchmark spanning eight recurring paper components, and
    \textbf{CS-50k}, a companion corpus for post-training.

    \item We develop a reproducible paper-to-QA construction pipeline that combines document
    parsing, retrieval-grounded question and answer generation, automatic quality control, and
    expert validation, enabling systematic construction of paper-derived QA resources from
    large paper collections.

    \item Extensive evaluations show that CS-4k distinguishes existing LLMs and exposes
    component-specific weaknesses, particularly on experimental settings and results.
    Post-training on CS-50k further yields substantial improvements, demonstrating the practical
    utility of structured, paper-derived QA supervision.
\end{itemize}

\begin{figure}[ht]
    \centering
    \includegraphics[width=1\textwidth]{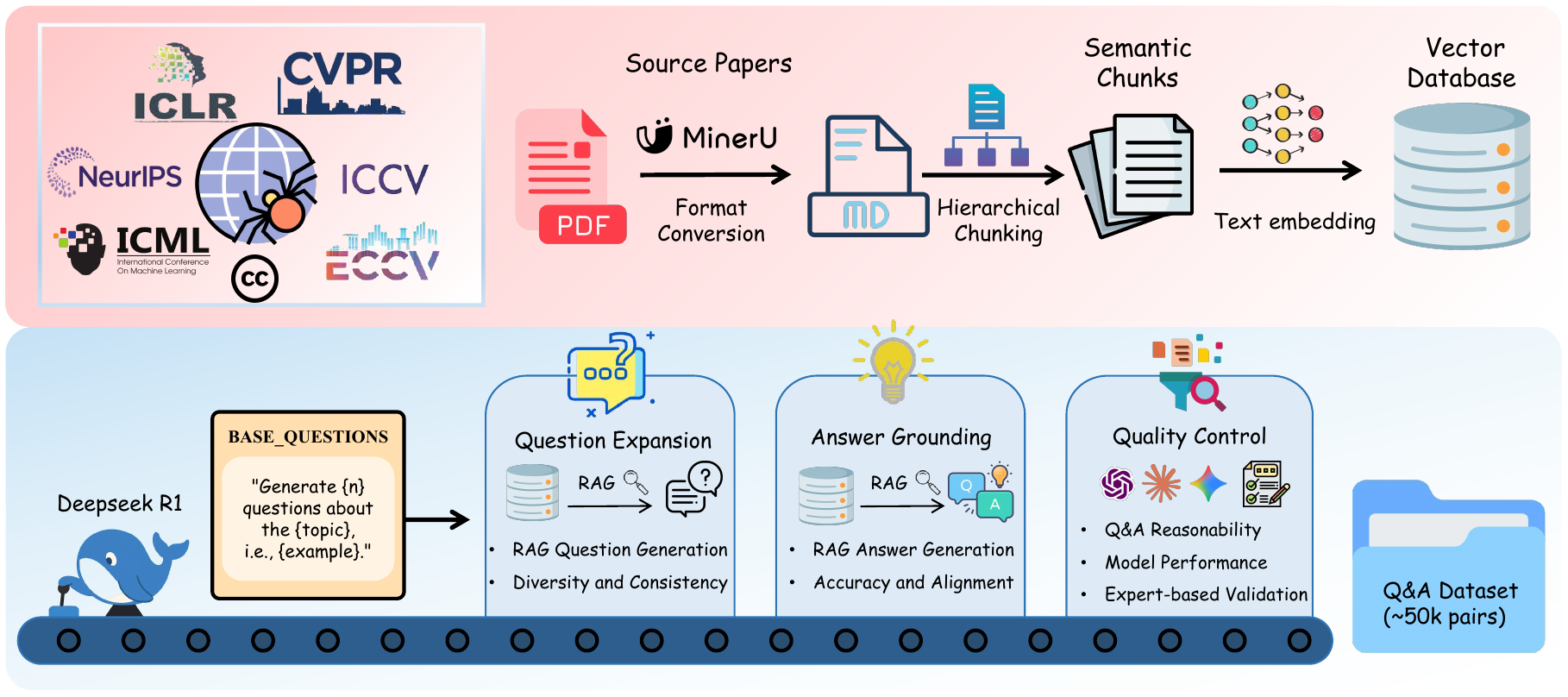}
    \caption{
    \textbf{Dataset construction pipeline. }
    \textbf{Top:} source papers from six AI/ML/CV conferences are converted, chunked, and embedded.
    \textbf{Bottom: }question expansion, answer grounding, two-stage automatic filtering, and
    sampled expert validation yield approximately 54k high-quality Q\&A pairs.}
    \label{fig:dataset-pipeline}
\end{figure}

\section{Related Work}
\subsection{Benchmarks for Scientific QA}
Recent advances in LLMs have led to the development of numerous benchmarks for evaluating their
capabilities across diverse domains and tasks. Early benchmarks in natural language processing (NLP)
were often built around relatively simple and domain-specific tasks, such as corpus annotation,
relation extraction, or fact-based question answering
\citep{kim2003genia,luan2018multi,jin2019pubmedqa}. As LLMs
progressed, benchmark design gradually shifted toward broader and more challenging evaluations,
covering both general-purpose reasoning and discipline-oriented examinations. General-purpose
evaluations such as BIG-Bench \citep{srivastava2023BIGbench} and HELM \citep{helm} provide broad
coverage across domains, while exam-style datasets including MMLU \citep{mmlu}, GPQA
\citep{rein2024gpqa}, SuperGPQA \citep{du2025supergpqa}, and Humanity's Last Exam (HLE)
\citep{phan2025HLE} test knowledge and reasoning at different academic levels. Scientific reasoning
benchmarks such as CURIE \citep{cui2025curie} and SFE \citep{zhou2025SFE} further emphasize
multi-step problem solving and multimodal understanding in scientific contexts. Recent initiatives
like ResearchBench \citep{liu2025researchbench} and MOOSE-Chem2 \citep{yang2025moose} examine
hypothesis generation and fine-grained scientific reasoning, while systems such as Agent Laboratory
\citep{schmidgall2025agent} and the AI Scientist \citep{lu2024ai} push toward autonomous research
workflows.
In contrast, CS-54k focuses on converting AI research papers into structured QA resources, so that
models can be evaluated and adapted on questions grounded in paper components such as motivation,
methods, experimental settings, and results.

\subsection{LLM Alignment}

LLM alignment aims to ensure that model outputs align with human
intentions.\citep{bai2022constitutional,rafailov2023direct,dong2023raft}.
Early attempts relied on supervised fine-tuning (SFT) on limited, human-annotated data, which
improved instruction-following but was constrained by the coverage and quality of the data
\citep{weifinetuned, wang2023self}.
InstructGPT \citep{ouyang2022training} then introduced the now-standard RLHF paradigm--combining SFT
with a reward model and PPO--to achieve stronger alignment.
More recently, GRPO \citep{shao2024deepseekmath} has emerged as a more efficient alternative,
streamlining the RLHF pipeline by eliminating the value model and stabilizing optimization. A
persistent challenge in this line of work is \emph{reward hacking} \citep{gao2023scaling}, where
models exploit imperfections in reward models, leading to degraded true alignment. Mitigation
strategies such as ensemble rewards and constrained objectives have been proposed
\citep{costereward,moskovitzconfronting}. While prior alignment efforts emphasized dialogue and
safety, here we study SFT and GRPO as post-training methods for paper-derived QA.

\begin{figure}[!t]
  \centering
  \begin{minipage}{0.42\linewidth}
    \centering
    \includegraphics[width=\linewidth]{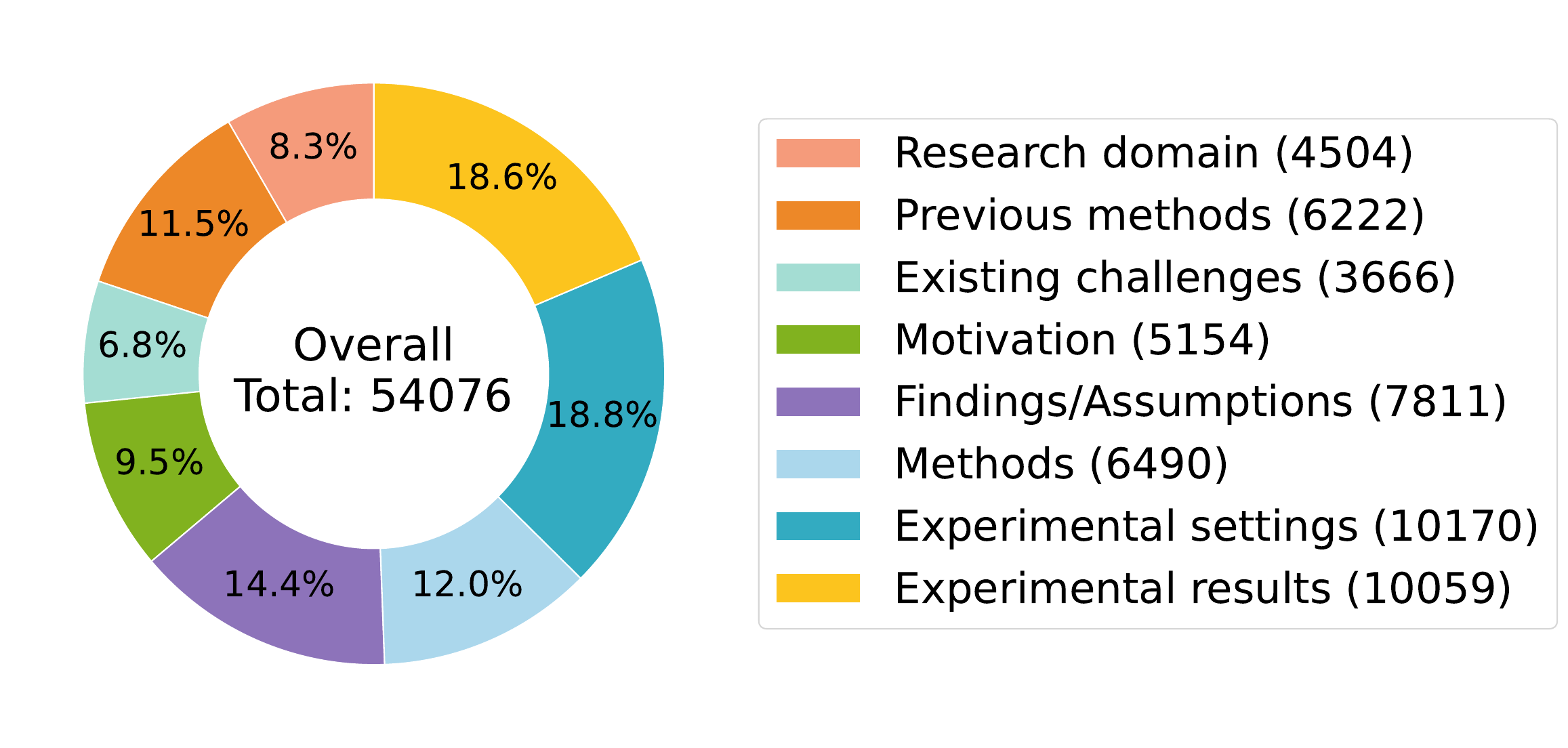}
    \centerline{\small (a) Category distribution}
  \end{minipage}\hfill
  \begin{minipage}{0.28\linewidth}
    \centering
    \includegraphics[width=\linewidth]{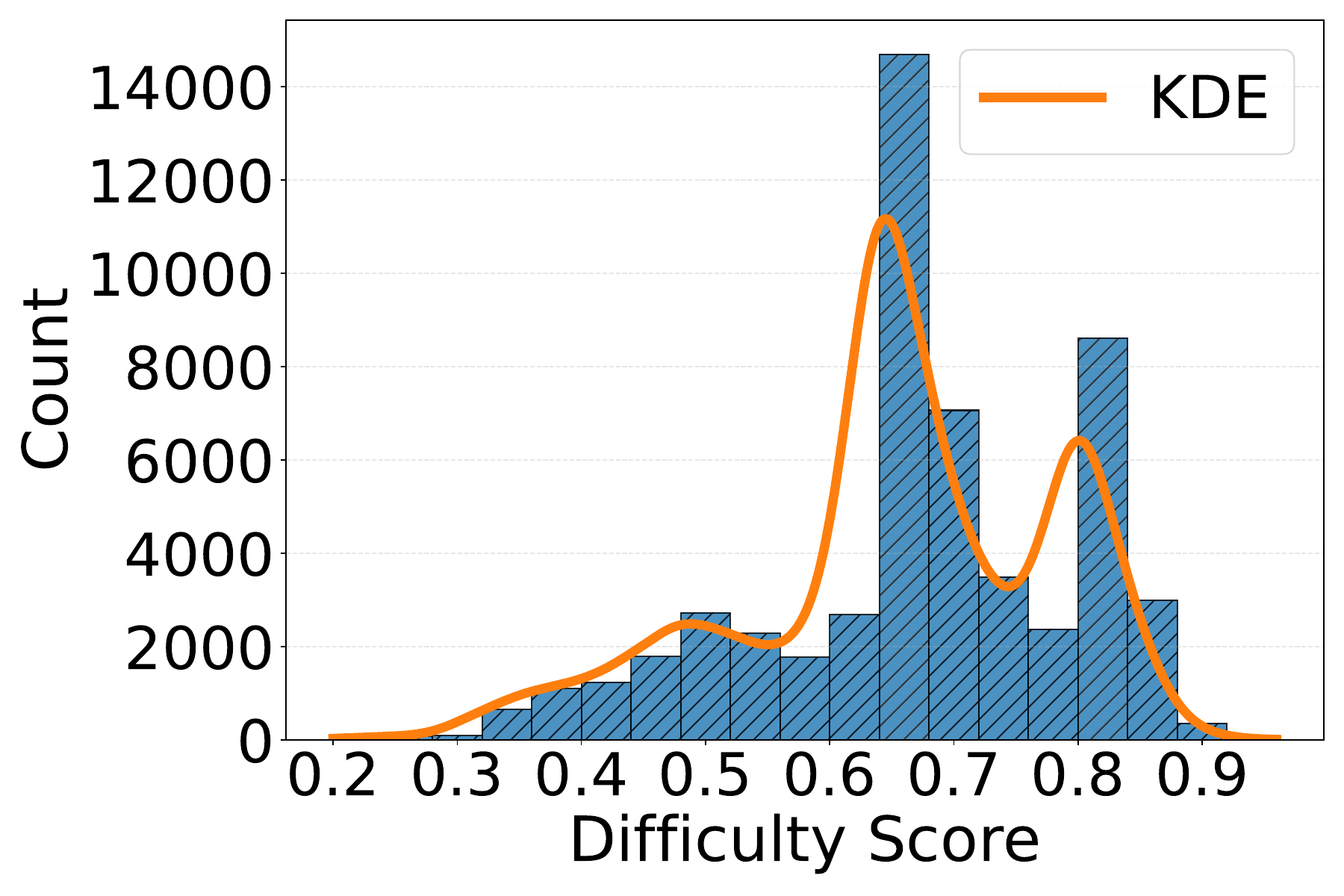}
    \centerline{\small (b) Difficulty distribution}
  \end{minipage}\hfill
  \begin{minipage}{0.28\linewidth}
    \centering
    \includegraphics[width=\linewidth]{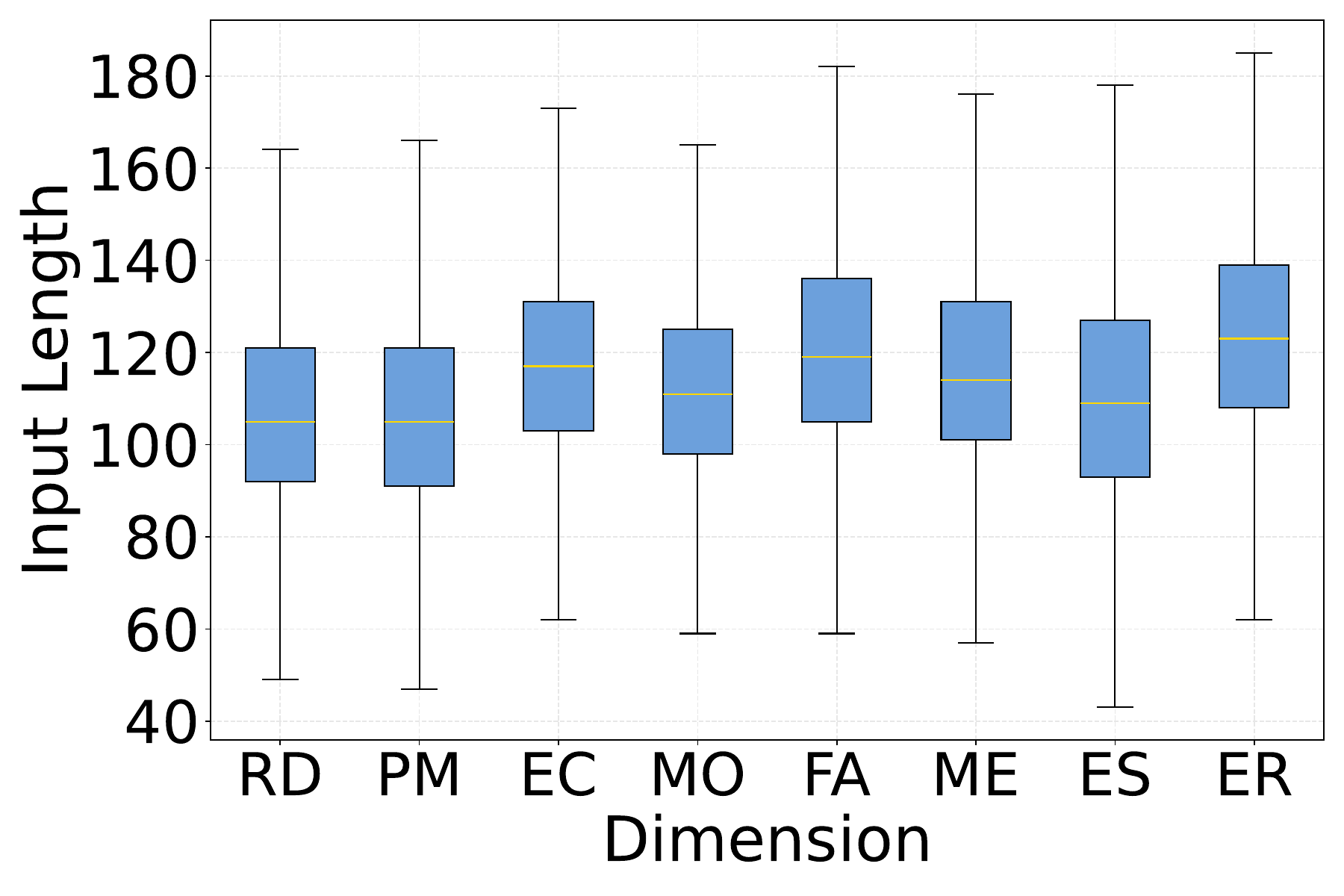}
    \centerline{\small (c) Input length distribution}
  \end{minipage}
  \caption{Overall dataset statistics for CS-4k/50k combined.
  (a) Category coverage across 8 paper components.
  (b) the distribution of difficulty scores estimated by an LLM-based scorer (histogram with KDE
  fit).
  (c) Input length variation across categories.}
  \label{fig:overall-stats}
\end{figure}

\section{Data Construction}
To construct CS-54k, we develop a systematic pipeline that converts large collections of
AI research papers into structured Q\&A pairs supported by source evidence.
The pipeline covers paper collection, document processing, question and answer generation,
and multi-stage quality control, as illustrated in Figure~\ref{fig:dataset-pipeline}.
\subsection{Data Collection}

We begin by collecting approximately 66,000 papers from arXiv, restricted to six AI, machine
learning, and computer vision conferences: ICML, ICLR, NeurIPS, ICCV, ECCV, and CVPR.
Among these, only papers released under Creative Commons open-access licenses are retained,
resulting in a final corpus of 14,474 papers and ensuring both legal compliance and reproducibility.

Each paper is converted from PDF format into Markdown using the
MinerU\citep{wang2024mineruopensourcesolutionprecise,he2024opendatalab} toolkit. This conversion
facilitates text-level operations such as segmentation and semantic analysis. To preserve contextual
integrity, we apply hierarchical chunking, dividing each paper into semantically coherent segments.

We embed each chunk using the \emph{nomic-embed-text-v1.5}\citep{nussbaum2024nomic} model and store
the representations in a vector database. This enables efficient retrieval of relevant passages
during question generation and ensures that the dataset construction process remains grounded in the
original paper content.
For each paper, we organize its content into eight recurring components of AI research papers:
\emph{research domain}, \emph{previous methods}, \emph{existing challenges}, \emph{motivation},
\emph{findings or assumptions}, \emph{methods}, \emph{experimental settings}, and \emph{experimental
results}.
These categories capture common semantic dimensions of AI research papers and guide downstream
question construction.

\subsection{Data Generation}
Following the segmentation and annotation of papers into topic categories, we proceed to generate
Q\&A pairs. Instead of generating questions heuristically, we adopt a RAG
\citep{lewis2020retrieval,gao2023retrieval,zhao2024retrieval,fan2024survey} pipeline.
This approach allows us to anchor both the questions and answers directly to the original paper
content, thereby minimizing hallucination and ensuring factual grounding.

For each topic class, we design a \emph{base question template} to guide the question generation
process. The template is:
"Generate \{n\} questions about the \{topic\}, i.e., \{example\}."
This template provides controlled guidance on the type of questions expected within each category,
ensuring consistency and relevance across the dataset. Here, \{topic\} corresponds to the eight
paper components listed above, and \{example\} is instantiated using their explanations.

Based on these templates, question construction is conducted in two stages:

\begin{enumerate}

    \item \textbf{Template Expansion with Evidence Retrieval.} First, we apply the
    \texttt{QUESTION\_AUG\_PROMPT}, listed in
    Appendix~\ref{app:data-generation-prompts}, to expand the template into several diverse drafts.
    Then, each draft is grounded into a fully specified question through a RAG process, using the
    \texttt{QUESTION\_GEN\_PROMPT} in Appendix~\ref{app:data-generation-prompts} together with
    relevant chunks from the vector database. This step ensures a variety of questions while
    maintaining alignment with the original paper content.

    \item \textbf{Answer Grounding.} Each candidate question is paired with an answer generated
    through the \texttt{ANSWER\_GEN\_PROMPT} in
    Appendix~\ref{app:data-generation-prompts}. Each draft retrieves its own chunk-level evidence
    independently, and retrieved passages are not intentionally shared across questions, reducing
    exact evidence reuse.
     RAG plays a key role in ensuring that the generated answers are grounded in the paper's
     content, aligning both questions and answers in a cohesive manner.

\end{enumerate}

Through this two-stage RAG-driven process, we obtain approximately \textbf{600,000 preliminary Q\&A
pairs}, covering a broad range of aspects from background knowledge to methodological details and
experimental findings.

\subsection{Quality Control}
To ensure the reliability of the dataset, we use two-stage automatic filtering followed
by sampled expert validation:

\begin{itemize}

    \item \textbf{Q\&A Coarse Filtering.} We employ a model-based evaluation (DeepSeek-R1) to
    automatically assess Q\&A pairs and filter out unreasonable or low-quality samples. The
    evaluation relies on a dedicated answer-evaluation prompt
    (Appendix~\ref{app:quality-control-prompts}) that guides the model in checking semantic coherence
    and factual consistency. This process reduces the dataset to around \textbf{100k high-quality
    Q\&A pairs}, which are then standardized and reformatted for consistency.

    \item \textbf{Fine-Grained Filtering.} For each question, we obtain three answers from
    \texttt{GPT-4-mini}, \texttt{Gemini 2.5 Flash}, and \texttt{Claude 3.5 Haiku}.
    We then use DeepSeek-R1 with a dedicated anomaly-detection prompt
    (Appendix~\ref{app:quality-control-prompts}) to evaluate the three answers.
    R1 checks two signals: strong disagreement across the answers, which indicates ambiguity, and
    nonsensical or generic responses, which indicate low information. Pairs that exhibit these
    abnormal patterns are removed automatically.

    \item \textbf{Expert-based Validation.}
    To validate the automatically filtered dataset, we conduct an expert-based
    evaluation on a stratified subset.
    We uniformly sample \textbf{160} Q\&A pairs across all categories to ensure balanced coverage.
    Each instance is independently reviewed by two expert annotators using binary (0/1) judgments.
    The expert evaluation yields an aggregated pass rate of 92.2\%, together with an
    inter-annotator agreement of Cohen's $\kappa = 0.87$, indicating strong agreement.
    Detailed results and annotation guidelines are provided in the
    Appendix~\ref{app:expert-validation}


\end{itemize}

\subsection{Data Quality and Analysis}

A single research paper may support questions across multiple components of the research process.
We therefore treat the QA instance as the basic unit of data construction and evaluation.
The corpus is partitioned into two complementary subsets:
\textbf{CS-50k} for post-training and \textbf{CS-4k} for evaluation.
After automatic filtering, 4,050 high-quality samples are selected through
distribution-preserving sampling to form CS-4k, while the remaining 50,026 samples form CS-50k.

The two subsets contain disjoint QA instances, and each question is grounded using independently
retrieved evidence from its source paper.
Because different questions derived from the same paper may appear in different subsets, the
post-training study evaluates the utility of CS-50k for improving performance on unseen QA
instances within a shared paper collection, rather than transfer to entirely unseen papers.
This QA-level design preserves the multi-component structure of research papers while supporting
both existing-model benchmarking and training-corpus analysis.

During splitting, we preserve the overall distributions of paper components, difficulty
levels, and input lengths to ensure balanced coverage across subsets.
The resulting corpus contains \textbf{54,076 high-quality QA pairs} spanning eight recurring
paper components:
\emph{research domain, previous methods, existing challenges, motivation,
findings/assumptions, methods, experimental settings, and experimental results}
(Figure~\ref{fig:overall-stats}a).
Each question is assigned a continuous difficulty score estimated by an LLM-based assessor from the
\texttt{Data-Juicer} library \citep{djv1,djv2}; the difficulty-scoring prompt is provided in
Appendix~\ref{app:quality-control-prompts}.
Input-length distributions show moderate variation across components without substantial skew
(Figure~\ref{fig:overall-stats}c).

\section{Evaluation of Existing LLMs on CS-4k}
\label{exp}

In this section, we evaluate existing LLMs on \textbf{CS-4k}, a benchmark consisting of
paper-derived questions with reference answers grounded in the source papers.
We primarily evaluate models in a closed-book setting, where the model receives no paper context
beyond the question.
We also include a complementary source-paper-access analysis for selected models.
Table~\ref{tab:results} reports results for both existing LLMs and Qwen variants post-trained
on CS-50k.

\subsection{Experimental Setup}
We evaluate a broad range of LLMs on CS-4k. For reasoning models,
we evaluate: DeepSeek-R1\citep{guo2025deepseek},
GPT-5\citep{gpt5},
o3,
o4-mini\citep{gpto3-o4mini},
o3-mini\citep{gpto3mini},
o1-mini\citep{gpto1mini},
Gemini 2.5 Pro,
Gemini 2.5 Flash\citep{team2023gemini},
Claude 4 Opus,
Claude 4 Sonnet,
Claude 3.7 Sonnet\citep{anthropic2023},
Grok4,
Grok3 mini fast,
Grok3 mini\citep{grok2023},
Qwen3-30B-A3B-Thinking-2507,
Qwen3-235B-A22B-Thinking-2507\citep{qwen3technicalreport}.
For chat models, we evaluate:
GPT-4.1,
GPT-4.1-mini\citep{gpt4.1-and-mini},
GPT-4o\citep{gpt4orelease},
GPT-4o-mini\citep{gpt4omini},
Gemini 2.0 Flash\citep{team2023gemini},
Claude 3.5 Haiku\citep{anthropic2023},
Grok3\citep{grok2023},
Llama-3.3-70B-Instruct\citep{llama3modelcard},
Qwen3-30B-A3B-Instruct-2507,
Qwen3-235B-A22B-Instruct-2507\citep{qwen3technicalreport}

\paragraph{Metrics.}
When evaluating models, we configure all temperatures to 0 for reduced randomness and employ a
unified zero-shot prompt template across all tasks.
To assess answer quality, we adopt an \textbf{LLM-as-a-Judge} paradigm, using
\texttt{gpt-4.1-2025-04-14} as the primary evaluation model.
The scoring prompt, provided in Appendix~\ref{app:quality-control-prompts}, instructs the judge to
assign an integer score from 0 to 10 based on semantic and technical alignment with the reference answer.
Each model prediction is evaluated along eight predefined research-paper categories, and an
\textbf{Overall} score is computed by averaging across categories.

To assess the reliability of the LLM-as-a-Judge evaluation, we conduct a human verification study
using a ranking-based protocol.
Across three representative judges, the judge-induced rankings show strong agreement with human
rankings, with mean Kendall's $\tau$ values of 0.81 (GPT-4.1), 0.79 (DeepSeek-R1), and 0.76 (Gemini
2.5).
This result suggests that the LLM-as-a-Judge framework is broadly aligned with human comparative
judgments in this setting.
Detailed evaluation protocols and results are provided in Appendix~\ref{sec:human-validation}.

\subsection{Model Evaluation Results}
We evaluate diverse LLMs on \textbf{CS-4k} using an LLM-as-a-Judge metric. Table~\ref{tab:results}
shows that CS-4k reveals fine-grained performance differences across existing models.

\begin{table}[!t]
\centering
\tablestyle{5pt}{1.1}
\resizebox{\textwidth}{!}{%
\begin{tabular}{p{4.8cm}<{\raggedright\arraybackslash}*{9}{p{1.6cm}<{\centering\arraybackslash}}}
\toprule
\textbf{Model} & \textbf{\shortstack{Research\\domain}} & \textbf{\shortstack{Previous\\methods}} & \textbf{\shortstack{Existing\\challenges}} & \textbf{Motivation} & \textbf{\shortstack{Findings/\\assump.}} & \textbf{Methods} & \textbf{\shortstack{Exp.\\settings}} & \textbf{\shortstack{Exp.\\results}} & \textbf{Overall} \\
\midrule
\rowcolor{color11}
\multicolumn{10}{c}{\textbf{\textit{Reasoning Models}}}\\
\midrule
\rowcolor{color12}
DeepSeek-R1 & 55.93 & 53.80 & 57.08 & 61.09 & 57.16 & 55.91 & 48.04 & 54.42 & 54.66 \\
\rowcolor{color12}
GPT-5 & 59.08 & 58.54 & 66.75 & 65.05 & 60.79 & 63.19 & 40.10 & 51.39 & 55.83 \\
\rowcolor{color12}
o3 & \textbf{71.66} & \textbf{71.05} & \textbf{78.83} & \textbf{77.56} & \textbf{75.71} & \textbf{74.40} & \textbf{59.57} & \textbf{72.07} & \textbf{71.35} \\
\rowcolor{color12}
o4-mini & 65.04 & 64.36 & 69.34 & 72.38 & 68.84 & 67.61 & 54.25 & 65.68 & 64.90 \\
\rowcolor{color12}
o3-mini & 56.26 & 53.95 & 54.49 & 59.95 & 57.61 & 56.73 & 48.58 & 56.50 & 55.08 \\
\rowcolor{color12}
o1-mini & 42.88 & 40.71 & 41.79 & 46.94 & 44.96 & 44.32 & 34.33 & 43.54 & 41.93 \\
\rowcolor{color12}
Gemini 2.5 Pro & 49.58 & 48.28 & 48.28 & 56.76 & 53.32 & 51.56 & 40.46 & 52.00 & 49.54 \\
\rowcolor{color12}
Gemini 2.5 Flash & 45.67 & 42.68 & 45.40 & 51.74 & 47.16 & 47.43 & 36.36 & 43.58 & 44.17 \\
\rowcolor{color12}
Gemini 2.5 Flash Search & 52.94 & 47.77 & 47.70 & 54.77 & 45.54 & 48.42 & 32.65 & 42.94 & 44.87 \\
\rowcolor{color12}
Claude 4 Opus & 45.61 & 39.70 & 43.43 & 50.21 & 44.21 & 45.51 & 30.09 & 38.06 & 40.68 \\
\rowcolor{color12}
Claude 4 Sonnet & 43.86 & 40.04 & 44.01 & 49.04 & 43.42 & 44.84 & 31.21 & 37.86 & 40.48 \\
\rowcolor{color12}
Claude 3.7 Sonnet & 41.63 & 36.46 & 39.89 & 48.08 & 41.79 & 41.81 & 29.55 & 33.50 & 37.79 \\
\rowcolor{color12}
Grok 4 & 45.99 & 44.79 & 47.08 & 50.91 & 47.26 & 47.06 & 36.46 & 45.03 & 44.73 \\
\rowcolor{color12}
Grok 3 mini fast & 44.45 & 42.25 & 43.76 & 49.72 & 45.16 & 45.78 & 34.93 & 42.45 & 42.75 \\
\rowcolor{color12}
Grok 3 mini & 44.78 & 41.31 & 43.21 & 49.66 & 44.56 & 44.90 & 38.02 & 42.15 & 42.96 \\
\rowcolor{color12}
Qwen3-30B-A3B-Thinking-2507 & 43.32 & 41.91 & 52.77 & 57.36 & 46.68 & 45.12 & 34.53 & 46.38 & 44.75 \\
\rowcolor{color12}
Qwen3-235B-A22B-Thinking-2507 & 51.63 & 47.27 & 59.45 & 61.84 & 51.64 & 52.14 & 35.14 & 48.21 & 48.96 \\

\midrule
\rowcolor{color21}
\multicolumn{10}{c}{\textbf{\textit{Chat Models}}}\\
\midrule
\rowcolor{color22}
GPT-4.1 & \textbf{51.25} & \textbf{49.01} & \textbf{49.56} & \textbf{54.15} & \textbf{53.33} & \textbf{52.74} & 42.11 & \textbf{52.52} & \textbf{50.15} \\
\rowcolor{color22}
GPT-4.1-mini & 49.73 & 47.55 & 47.08 & 53.11 & 51.40 & 51.48 & \textbf{42.20} & 50.78 & 48.85 \\
\rowcolor{color22}
GPT-4o & 42.37 & 39.72 & 38.32 & 45.39 & 42.77 & 42.26 & 34.30 & 41.47 & 40.44 \\
\rowcolor{color22}
GPT-4o-mini & 40.95 & 38.28 & 37.45 & 43.42 & 40.63 & 40.78 & 32.97 & 41.25 & 39.13 \\
\rowcolor{color22}
GPT-4o mini Search Preview & 39.02 & 30.97 & 33.94 & 38.06 & 33.06 & 33.81 & 17.73 & 27.33 & 27.33 \\
\rowcolor{color22}
Gemini 2.0 Flash & 40.33 & 37.79 & 39.67 & 44.15 & 40.50 & 42.18 & 28.03 & 34.96 & 37.29 \\
\rowcolor{color22}
Gemini 2.0 Flash Search & 31.34 & 30.30 & 28.10 & 37.12 & 26.24 & 29.88 & 13.41 & 20.84 & 25.31 \\
\rowcolor{color22}
Claude-3.5 & 28.81 & 25.19 & 30.77 & 23.70 & 20.53 & 28.05 & 16.77 & 16.10 & 22.12 \\
\rowcolor{color22}
\rowcolor{color22}
Llama-3.3-70B-Instruct & 39.41 & 36.39 & 35.36 & 41.84 & 39.20 & 39.09 & 30.75 & 37.86 & 37.03 \\
\rowcolor{color22}
Qwen3-30B-A3B-Instruct-2507  & 44.90 & 43.09 & 44.85 & 50.03 & 47.49 & 45.88 & 36.72 & 45.70 & 44.28 \\
\rowcolor{color22}
Qwen3-235B-A22B-Instruct-2507 & 46.50 & 43.80 & 45.88 & 53.21 & 46.94 & 46.44 & 37.99 & 43.93 & 44.76 \\
\midrule
\rowcolor{color31}
\multicolumn{10}{c}{\textbf{\textit{CS-50k Post-Training Study}}}\\
\midrule
\rowcolor{color32}
Qwen2.5-7B-Instruct &38.87 &35.71 &35.55 &41.66 &37.71 &38.56 &30.21 &33.86 &35.78 \\
\rowcolor{color32}
Qwen2.5-7B-Instruct-SFT &42.55 &40.26 &43.83 &47.98 &47.45 &44.09 &35.22 &46.31 &43.11\\
\rowcolor{color32}
Qwen2.5-7B-Instruct-GRPO &48.96 &45.88 &46.53 &56.92 &57.06 &51.19 &\textbf{43.46} &56.31 &50.97\\
\rowcolor{color32}
Qwen2.5-7B-Instruct-GRPO-2RMs &\textbf{51.75} &\textbf{48.26} &\textbf{50.36} &\textbf{61.97} &\textbf{59.09} &\textbf{55.31} &42.94 &\textbf{57.63} &\textbf{53.15}\\
\bottomrule
\end{tabular}}
\caption{Dimension-wise evaluation results on CS-4k.
The first two blocks report evaluation of existing models.
The final block reports Qwen2.5-7B variants trained on CS-50k to evaluate CS-50k as supervision on
question-disjoint CS-4k samples drawn from the same paper collection.
\textbf{Bold} indicates the best result within each model group.}
\label{tab:results}
\end{table}

\paragraph{CS-4k reveals clear performance tiers across models.}
In the closed-book setting, CS-4k reveals clear performance differences among LLMs. o3 (71.35) and
o4-mini (64.90) rank well above mid-tier models such as GPT-5 (55.83) and DeepSeek-R1 (54.66), while
general chat models score lower (e.g., GPT-4.1 50.15, GPT-4o 40.44, Gemini 2.0 Flash 37.29). This
pattern appears across dimensions, showing that CS-4k can distinguish fine-grained performance gaps.
Figure~\ref{fig:cs4k_results} provides a compact visual summary of these overall trends.
Beyond absolute tiers, CS-4k also reveals differences within model families. In the Qwen3 family,
\emph{Thinking} variants score higher than their \emph{Instruct} counterparts at both 30B and 235B
scales.
This pattern is an existing-model benchmark observation under the CS-4k protocol, not a general claim
about reasoning ability.

\begin{figure}[!t]
    \centering
    \includegraphics[width=1\textwidth]{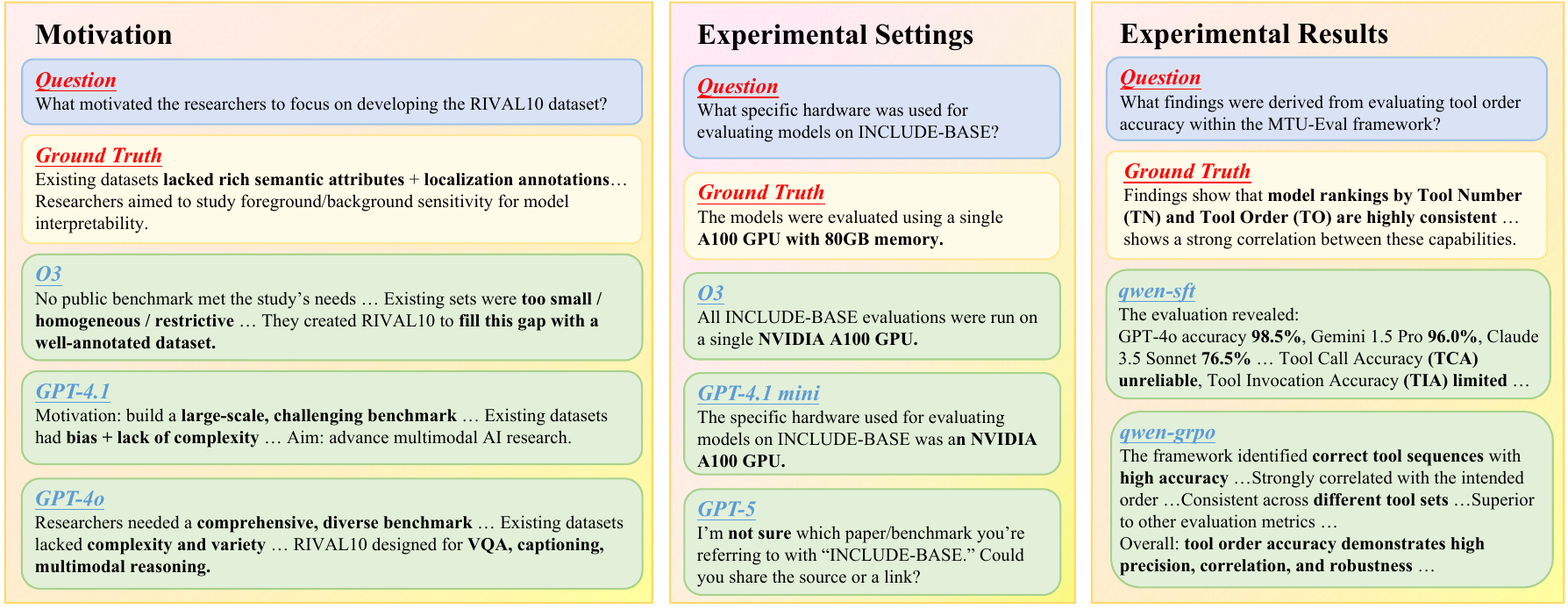}
    \caption{Case studies on CS-4k across different paper components.}
    \label{fig:case_study}
\end{figure}

\paragraph{CS-4k reveals component-specific performance gaps.}
Even the strongest models perform unevenly across paper components. For example, o3 achieves the
best results on most categories, including \emph{Existing Challenges} (78.83) and \emph{Motivation}
(77.56), but drops substantially on \emph{Experimental Settings} (59.57). This pattern is consistent
across models: they generally perform better on higher-level understanding tasks such as \emph{Existing
Challenges}, \emph{Motivation}, and \emph{Methods}, but remain weaker on technically specific
dimensions such as \emph{Experimental Settings} and \emph{Experimental Results}. These results
suggest that current LLMs are stronger at interpreting and summarizing research narratives than at
recovering precise experimental details. While they perform well on higher-level interpretation and
problem framing, they remain less reliable on tasks that require accurate reconstruction of
experimental protocols. Representative failure cases are shown in Figure~\ref{fig:case_study}.

\paragraph{Source-paper-access analysis.}\label{sec:Web-Augmented}

As a complementary analysis, we examine whether allowing selected models to consult
source papers improves performance on CS-4k.
We evaluate \texttt{GPT-4o-mini-search-preview}, \texttt{Gemini 2.0 Flash}, and
\texttt{Gemini 2.5 Flash} with their document-access capabilities enabled.
Under this setting, source-paper access does not consistently improve performance relative to the
closed-book evaluation.
Because the outcome jointly depends on document access, PDF parsing, evidence retrieval, and answer
generation, we treat this result as an exploratory system-level comparison rather than a controlled
evaluation of retrieval-based methods.
Detailed source-paper-access instructions are provided in Appendix~\ref{app:quality-control-prompts}.

\Needspace{10\baselineskip}
\section{Post-Training with CS-50k}
\label{sec:post-training}

We study whether CS-50k can serve as supervision for improving paper-derived QA.
Since CS-50k and CS-4k are question-disjoint subsets drawn from the same paper collection, we use
this experiment to evaluate the utility of CS-50k as supervision for paper-derived QA.
We select \textbf{Qwen2.5-7B-Instruct} as the backbone model and train it on the proposed
\textbf{CS-50k} dataset.
All experiments are conducted on 8$\times$NVIDIA A100-80G GPUs.
Detailed hyperparameter configurations are provided in Appendix~\ref{sec:training-hparams}.
For SFT, we employ the \textbf{LLaMA-Factory} framework\citep{zheng2024llamafactory}, training for
three epochs and selecting the checkpoint with the lowest validation loss.
For reinforcement optimization (GRPO), we adopt the \textbf{verl}
framework\citep{sheng2024hybridflow}, training for three epochs on top of the SFT checkpoint.
Based on this setup, we explore three strategies: SFT only, SFT+GRPO with a single reward model, and
SFT+GRPO with two reward models whose scalar rewards are averaged.

\begin{figure}[!ht]
    \begin{minipage}[b]{0.32\textwidth}
        \centering
        \includegraphics[width=\linewidth]{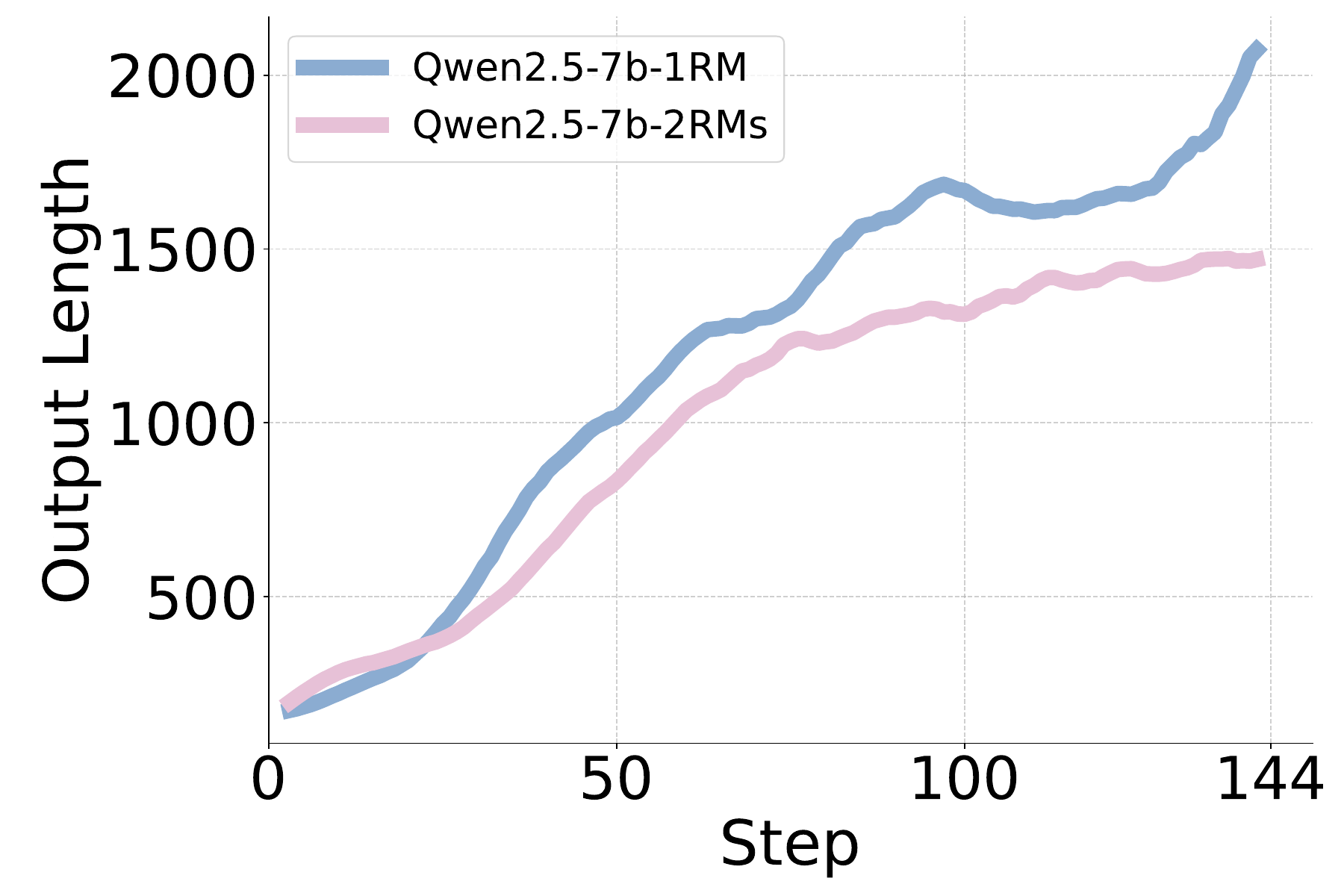}
    \end{minipage}
    \hfill
    \centering
    \begin{minipage}[b]{0.32\textwidth}
        \centering
        \includegraphics[width=\linewidth]{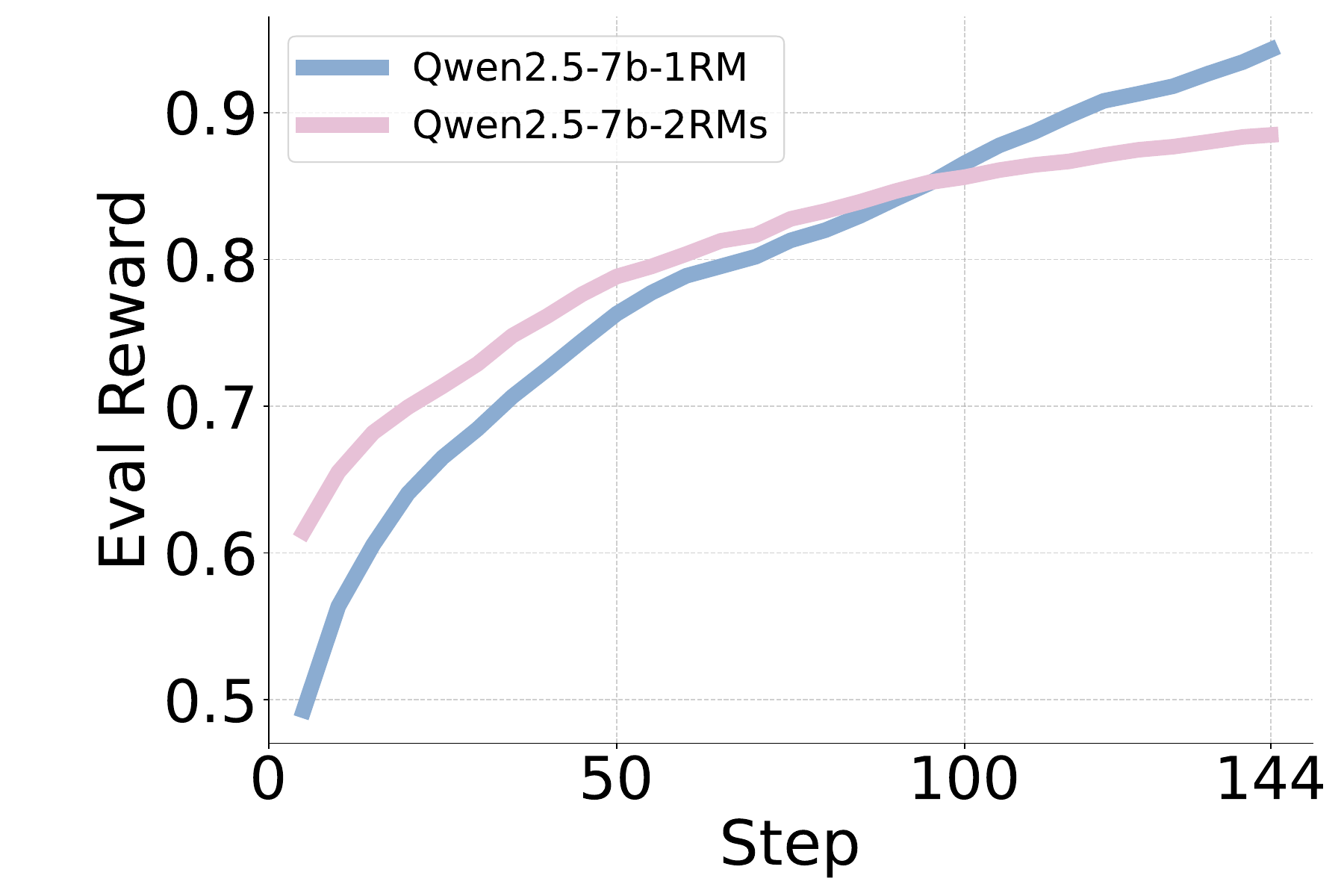}
    \end{minipage}
    \hfill
    \begin{minipage}[b]{0.32\textwidth}
        \centering
        \includegraphics[width=\linewidth]{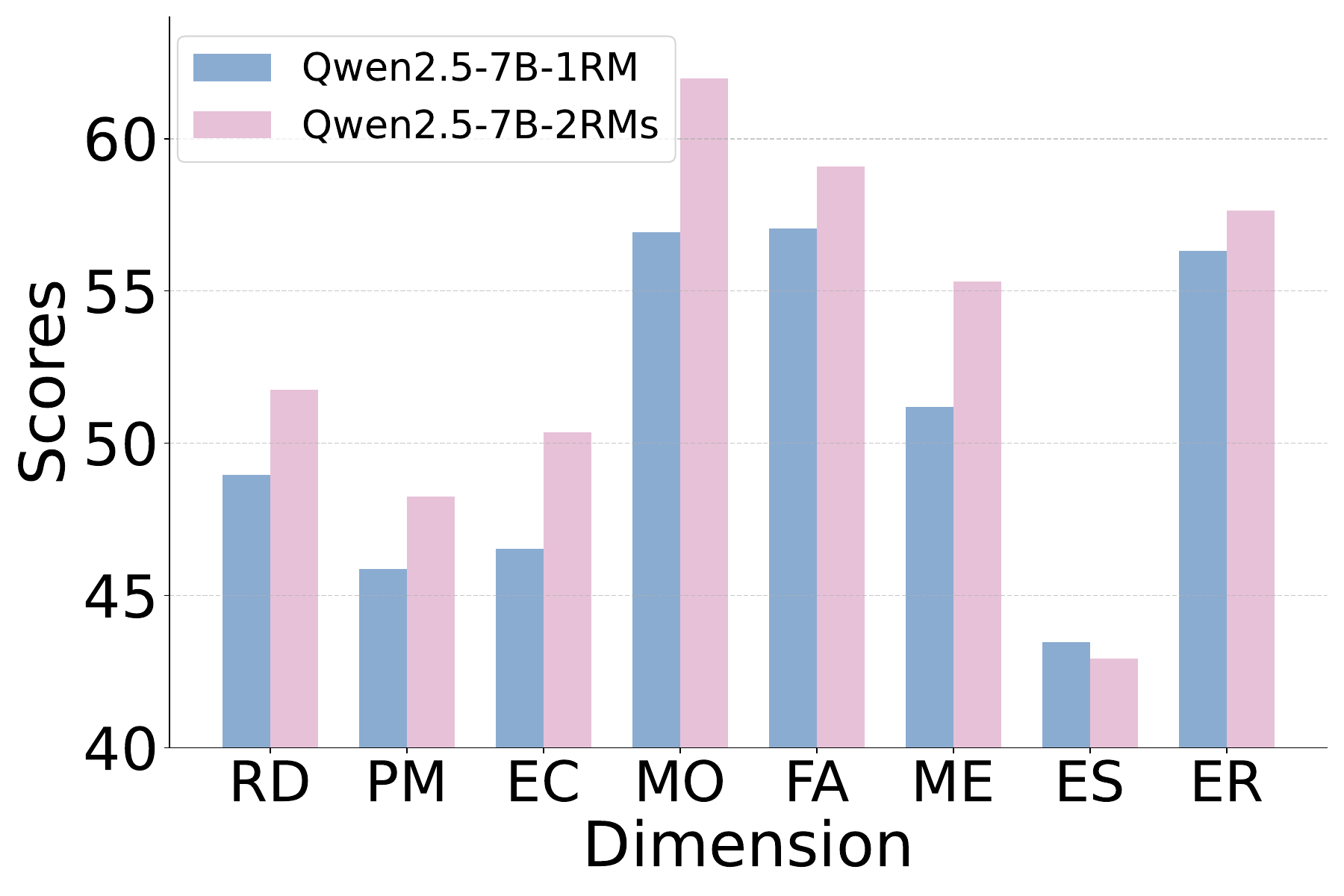}
    \end{minipage}
\caption{Training and evaluation statistics with single vs.\ dual reward models: output length
curves, evaluation reward curves, and CS-4k scores across paper components.}
    \label{fig:reward-hacking}
\end{figure}

\paragraph{CS-50k provides useful supervision for paper-derived QA.}
We next evaluate the impact of training on \textbf{CS-50k} by measuring model performance on
paper-derived questions in \textbf{CS-4k}.
Experiments on \texttt{Qwen2.5-7B-Instruct} provide evidence for the utility of CS-50k as a training
corpus.
The base model scores 35.78 overall, highlighting the limitations of general-purpose chat models on
paper-derived QA.
SFT on CS-50k raises the score to 43.11 (+7.33), indicating that structured paper-derived
supervision can improve performance on question-disjoint CS-4k samples.
Reinforcement optimization with GRPO further improves the overall score to 50.97
(+7.86 points over SFT), with the largest gains in \emph{Experimental Results} (+10.00 points)
and \emph{Findings/Assumptions} (+9.61 points).
\paragraph{Mitigating reward over-optimization with dual reward models.}
To reduce reward over-optimization, we adopt a dual reward model (2RM) setup with Qwen2.5-7B-Instruct and
Llama-3.1-8B-Instruct as parallel evaluators. This design yields a more stable optimization signal
and improves the overall CS-4k score to 53.15 in the CS-50k post-training study.
Figure~\ref{fig:reward-hacking} helps explain this improvement. In early training, 2RM achieves
higher rewards than 1RM, while in later stages 1RM shows inflated rewards accompanied by sharply
increased output length, suggesting over-optimization. By contrast, 2RM maintains more stable
rewards and controlled output length throughout training. When evaluated by GPT-4.1 on CS-4k, the 2RM model outperforms the
1RM model in seven of the eight components, with the largest gains in \emph{Motivation}
(+5.05 points) and \emph{Methods} (+4.12 points). Overall, dual-reward
optimization provides a more stable alignment signal in our experiments.

\paragraph{Case study.}\label{sec:case-study}
We present illustrative examples from three representative dimensions.

\emph{Motivation.}
For the RIVAL10 dataset, o3 correctly identified missing semantic attributes and localization
annotations as the motivation. GPT-4.1 noted dataset complexity, while GPT-4o produced only a vague
answer about building a ``comprehensive benchmark.'' This example shows that stronger models recover
paper-specific motivations, whereas other models may rely on generic narratives.

\emph{Experimental Settings.}
For hardware-configuration questions in INCLUDE-BASE, o3 and GPT-4.1-mini correctly identified the
use of a single NVIDIA A100 (80GB), whereas GPT-5 failed to answer. This highlights a common
weakness in recovering technical details important for reproducibility.

\emph{Experimental Results.}
On result-comparison questions, SFT-trained \texttt{Qwen} gives concise summaries but often
hallucinates, such as fabricating numbers or misattributing baselines in MTU-Eval. GRPO variants
still make minor numerical errors but better capture key findings. This contrast shows how different
training objectives shape the trade-off between brevity and factual fidelity.



\section{Conclusion and Discussion}


This paper introduces \textbf{CS-54k}, a paper-grounded QA corpus built from openly
licensed AI research papers and organized around recurring paper components.
Its reproducible paper-to-QA pipeline converts unstructured content into structured supervision
supported by source evidence for evaluation and post-training.
Experiments on \textbf{CS-4k} reveal model- and component-level differences, while
post-training with \textbf{CS-50k} improves \texttt{Qwen2.5-7B-Instruct}.
Together, CS-4k and CS-50k provide a unified resource for diagnosing and improving LLMs on
research paper QA.

Looking forward, the pipeline could be extended to incorporate figures and tables, supporting
multimodal paper-grounded QA and more comprehensive evaluation of multimodal research systems.

\bibliography{iclr2026_conference}
\bibliographystyle{iclr2026_conference}

\clearpage
\appendix
\raggedbottom
\setcounter{secnumdepth}{2}
\setlength{\floatsep}{6pt plus 1pt minus 1pt}
\setlength{\textfloatsep}{6pt plus 1pt minus 1pt}
\setlength{\intextsep}{6pt plus 1pt minus 1pt}
\section{Data Construction Prompts}
\label{app:prompts}
Our dataset is constructed through two types of prompts: \textbf{data generation prompts} and
\textbf{quality control prompts}. Each prompt is designed to improve grounding and diversity in the
generated data.
\subsection{Data Generation Prompts}
\label{app:data-generation-prompts}
We design three types of prompts to guide data generation.

\begin{promptbox}{Question Augmentation Prompt}
\label{tab:query-transformation}

You are a skilled linguistic assistant specializing in query transformation. When presented with a
user's query, your objective is to meticulously rephrase and enhance it by:

\begin{itemize}
  \item Utilizing synonyms, alternative phrasings, and varied sentence structures
  \item Adjusting grammatical form (e.g., converting commands to questions) while preserving intent
  \item Restructuring information flow without losing original components
  \item Improving clarity and natural fluency
  \item Maintaining all factual elements and nuanced meaning
\end{itemize}

Avoid adding new information or omitting key aspects. Prioritize conversational tone and logical
coherence in your revisions.

\medskip
\textbf{Format Requirements:}
\begin{itemize}
  \item Provide only the rewritten query
  \item No supplementary explanations or formatting
\end{itemize}

\medskip
\textbf{Example Transformation:}
\begin{description}
  \item[Original:] Give me 5 reasons of sleeping early
  \item[Response:] Can you list five advantages of maintaining an early bedtime schedule?
\end{description}

\textbf{Now transform this query while adhering to all guidelines:}

\begin{description}
  \item[Original:] \{input\}
  \item[Response:]
\end{description}

\end{promptbox}

\begin{promptbox}{Question Generation Prompt}
\label{tab:question-gen}

\textbf{Context information is below.}

\medskip
\texttt{---------------------} \\
\{context\} \\
\texttt{---------------------}

\medskip
\textbf{Objective}: Provide self-contained questions using \textbf{ONLY} the given context.
Formulate responses as if explaining to someone \textbf{UNAWARE} of this context in a JSON list.
When addressing domain-specific concepts:

\begin{itemize}
  \item Explicitly state the domain upfront
  \item Avoid all references to ``context'' or ``provided information''
  \item Assume zero prior knowledge
  \item Avoid vague or generalized references like ``this research'' or ``your research area.''
  Instead, directly address the specific subject matter or topic being discussed.
  \item For example, instead of asking ``What is the motivation of this work?'', you might consider
  framing the question more specifically, such as ``What motivated the researchers to focus on
  developing [specific subject]?'' or ``Why [specific subject] matters?''
\end{itemize}

\medskip
\textbf{Response Constraints}:
\begin{itemize}
  \item \textbf{FORBIDDEN} generalized phrases: [``provided context'', ``given context'', ``above
  information'', ``mentioned context'', ...]; replace them with specific subject matter
  \item For ambiguous terms (e.g., ``xxx''):
    \begin{itemize}
      \item First establish domain context
      \item Then provide explanation
    \end{itemize}
  \item Responses are required to be formatted as a JSON-parseable list: \texttt{["response1",
  "response2", ...]}
  \item Maintain consistent JSON syntax without explanatory text
  \item Absolute prohibition of:
    \begin{itemize}
      \item JSON/text hybrids
      \item Context-existence references
    \end{itemize}
  \item Ensure that the generated responses contain only a JSON list of several plain text questions
  without including codes, markdowns, or anything else.
\end{itemize}

\medskip
\textbf{Example Response (MUST be formatted as a JSON-parseable list):}

\texttt{["Question1", "Question2", "Question3", "Question4", "Question5"]}

\medskip
\textbf{Query:} \{input\} \\
\textbf{Answer:}

\end{promptbox}

\begin{promptbox}{Answer Generation Prompt}
\label{tab:answer-gen}

\textbf{Context information is below.}

\medskip
\texttt{---------------------} \\
\{context\} \\
\texttt{---------------------}

\medskip
Using \textbf{STRICTLY} the provided context text above, generate a concise, self-contained answer
to the query:

\medskip
\textbf{[PRIORITY 1: EVIDENCE-SUPPORTED ANSWERING]}
\begin{enumerate}
  \item \textbf{GROUNDING PRINCIPLE:}
  \begin{itemize}
    \item Preserve all factual details needed to answer the query
    \item Use only information supported by the retrieved context
    \item Do not copy unnecessary surrounding text from the source paper
    \item When multiple relevant pieces of evidence exist:
      \begin{enumerate}
        \item Combine them into a coherent standalone answer
        \item Keep the answer concise and specific to the query
        \item Replace context-specific references (e.g., ``this study'', ``our method'') with
        \texttt{[PROPOSAL]}
      \end{enumerate}
  \end{itemize}
\end{enumerate}

\medskip
\textbf{[PRIORITY 2: CONTEXTUAL SYNTHESIS]}
\begin{enumerate}
  \item If no single sentence fully answers but context contains relevant information:
  \begin{enumerate}
    \item Combine multiple context fragments into one self-contained answer using:
      \begin{itemize}
        \item Minimal transitional phrases
        \item Clear references to the relevant method, dataset, result, or assumption
      \end{itemize}
    \item Preserve factual meaning from the source material
    \item Remove non-essential surrounding text (e.g., ``interestingly'', ``as shown in Table 1'')
    \item Maintain factual integrity without adding unsupported interpretation
  \end{enumerate}
\end{enumerate}

\medskip
\textbf{[STRICT PROHIBITIONS]}
\begin{itemize}
  \item NEVER:
    \begin{itemize}
      \item Introduce information beyond context boundaries
      \item Add explanations not explicitly stated
      \item Paraphrase in ways that alter original meaning
      \item Speculation beyond context
      \item Include personal interpretations
    \end{itemize}
\end{itemize}

\medskip
\textbf{[FALLBACK INSTRUCTIONS]}
\begin{itemize}
  \item If context contains partial but insufficient information: State existing relevant facts
  \textbf{PRECISELY} in concise, natural wording
  \item Only output \texttt{"The context does not contain relevant information"} when:
    \begin{itemize}
      \item Zero contextual connection exists
      \item All potential answers require speculation
    \end{itemize}
\end{itemize}

\medskip
\textbf{[OUTPUT REQUIREMENTS]}
\begin{itemize}
  \item Use concise and natural wording while preserving all technical facts supported by the
  retrieved evidence
  \item Avoid unnecessary lexical copying from the source text
  \item Grammatical coherence for standalone understanding
\end{itemize}

\medskip
\textbf{Query:} \{input\} \\
\textbf{Answer:}

\end{promptbox}

\subsection{Quality Control Prompts}
\label{app:quality-control-prompts}
To evaluate the quality of generated data, we use several prompts for correctness checking,
difficulty assessment, and answer scoring.

\begin{promptbox2}{LLM Answer Evaluation Prompt}
\label{tab:Prompt for QA evaluation}
You will be given a question and an answer. Judge whether the answer is reasonable for the question.

\medskip
If the answer is relevant and correctly or appropriately answers the question, output 1.
If the answer is irrelevant, nonsensical, incorrect, or does not address the question, output 0.
Only output 0 or 1, and nothing else.

Now, judge the following:
\begin{description}
  \item[Question:] \{question\}
  \item[Answer:] \{answer\}
  \item[Output:]
\end{description}

\end{promptbox2}

\begin{promptbox2}{LLM Multi-Answer Anomaly Detection Prompt}
\label{tab:Prompt for Multi-Answer Evaluation}

You will be given a question, evidence, three LLM answers, and a reference answer.
Your task is to detect abnormal answer patterns.
Do not judge correctness.

\medskip
Two abnormal patterns are considered:

1. All three answers are unrelated to the reference answer (ambiguity or misalignment).
2. All three answers are nonsensical or generic (low information).

If any abnormal pattern appears, output 1.
If no abnormal pattern appears, output 0.
Only output 0 or 1, and nothing else.

Now, analyze the following:
\begin{description}
  \item[Question:] \{question\}
  \item[Evidence:] \{evidence\}
  \item[Reference Answer:] \{reference\_answer\}
  \item[Answer A:] \{answer\_a\}
  \item[Answer B:] \{answer\_b\}
  \item[Answer C:] \{answer\_c\}
  \item[Output:]
\end{description}

\end{promptbox2}

\begin{promptbox2}{LLM Difficulty Score Prompt}
\label{tab:llm-difficulty}

You are an expert pedagogical evaluator for LLM training data. Analyze each data sample through
multiple difficulty lenses and provide calibrated scores with detailed reasoning. Follow these
guidelines:

\medskip
\textbf{1. Evaluation Dimensions}
Rate each dimension (1--5 scale: 1=Novice-friendly, 3=Intermediate, 5=Expert-level):
\begin{itemize}
  \item Linguistic Complexity: Vocabulary sophistication \& syntactic structures
  \item Conceptual Depth: Abstraction level \& theoretical requirements
  \item Prior Knowledge: Required domain-specific understanding
  \item Step Complexity: Problem-solving steps needed
  \item Ambiguity: Multiple valid interpretations
\end{itemize}

\medskip
\textbf{2. Output Format}

\begin{lstlisting}[breaklines=true]
json
{
  "dimension_scores": {
    "linguistic_complexity": ,
    "conceptual_depth": ,
    "prior_knowledge": ,
    "step_complexity": ,
    "ambiguity":
  },
  "flags": ["multistep_reasoning",
            "cultural_context", ...],
  "rationale":
  "Technical analysis of challenge sources"
}
\end{lstlisting}

\medskip
\textbf{3. Special Instructions}
\begin{itemize}
  \item Differentiate intrinsic vs. extrinsic difficulty factors
  \item Account for varying cultural/educational backgrounds
  \item Mark samples requiring cross-domain knowledge synthesis
  \item Consider temporal aspects for time-sensitive subjects
  \item Flag ambiguous samples needing difficulty bracketing
  \item Response a JSON dict
\end{itemize}

\medskip
\textbf{Example Response:}

\begin{lstlisting}[breaklines=true]
json
{
  "dimension_scores": {
    "linguistic_complexity": 3,
    "conceptual_depth": 5,
    "prior_knowledge": 4,
    "step_complexity": 4,
    "ambiguity": 5
  },
  "flags": ["nonlinear_reasoning", "semantic_ambiguity"],
  "rationale": "High conceptual difficulty due to multi-layered
  metaphor interpretation requiring philosophy background.Moderate
  linguistic complexity offset by implicit cultural references."
}
\end{lstlisting}

\end{promptbox2}

\begin{promptbox2}{Evaluation prompt}
\label{tab:Evaluation prompt}

You are a grading expert. Judge whether the final answers given by the candidates below are
consistent with the reference answers, i.e., whether the candidates answered correctly. Provide a
single \textbf{integer score from 0 to 10}.

\vspace{0.5em}
\textbf{Scoring Guide:}

\begin{list}{}{%
  \setlength{\labelwidth}{2.8em}%
  \setlength{\leftmargin}{3.3em}%
  \setlength{\labelsep}{0.4em}%
  \setlength{\itemsep}{0.2em}%
  \setlength{\topsep}{0.2em}%
}
  \item[9--10:] Fully correct, detailed, and clearly structured.
  The prediction matches the reference not only in meaning, but also in technical detail,
  terminology, and step-by-step structure. Only assign this score if the answer is nearly
  indistinguishable from the reference.
  \item[7--8:] Covers all key points and includes all key technical details, but may have slight
  simplifications, structural looseness, or phrasing issues. The prediction is still technically
  sound and semantically aligned.
  \item[5--6:] Covers nearly all major technical points with acceptable clarity and structure. The
  prediction may slightly oversimplify or omit minor supporting details, but the core reasoning is
  intact.
  \item[3--4:] Incomplete or partially correct. Misses one or more core concepts or steps, or is
  expressed in a vague, confusing, or disorganized way.
  \item[1--2:] \textbf{Weak or flawed answer.} Contains major factual errors, misunderstands the
  question, or mixes unrelated ideas. May reference a few relevant terms, but lacks meaningful
  explanation or structure.
  \item[0:] \textbf{Fundamentally incorrect or completely off-topic.} No meaningful alignment with
  the reference content, or content is nonsensical or missing.
\end{list}

\vspace{0.5em}
\textbf{Output Format:}

\begin{itemize}
    \item Respond with a \textbf{single integer from 0 to 10}.
    \item Do not include any explanation or additional text.
\end{itemize}

\vspace{0.5em}
\textbf{Special Instructions:}
\begin{itemize}
  \item The model prediction may contain the reasoning process; you should spot the final answer
  from it.
  \item Do not re-answer the question yourself.
  \item Assign a high score only if the prediction matches the answer \textbf{semantically and
  technically}, considering variations in format.
  \item Deduct points for missing key technical details or excessive generalizations. Even if the
  tone is correct, factual or structural omissions should lead to a reduced score.
  \item Be strict with general or vague answers: If the prediction only provides a high-level
  overview but omits key technical details, steps, or quantitative findings present in the
  reference, score \textbf{no higher than 6}.
  \item Ignore minor differences in formatting, capitalization, or spacing.
\end{itemize}

\vspace{0.5em}
\textbf{Example Response:}
7

\vspace{0.5em}
\textbf{Now start your task.}

\medskip

\begin{description}
  \item[Question:] \{question\}
  \item[Reference Answer:] \{answer\}
  \item[Model Prediction:] \{prediction\}
\end{description}

\end{promptbox2}

\begin{promptbox2}{Web-augmented Question Formulation Prompt}
\label{tab:Web-augmented question formulation prompt}

Reference paper link: \{url\}

Use only the content from the paper at this URL.
Do not perform open-ended web search or consult external sources.
Ground your answer strictly in the referenced paper.

Please answer the following question: \{question\}

\end{promptbox2}

\section{Expert-based Validation Details}
\label{app:expert-validation}

We validate the constructed Q\&A pairs using \textbf{160} stratified samples from CS-54k,
with 20 pairs per paper component.
Each question-answer-paper triple is independently reviewed by \textbf{two expert annotators
with relevant research backgrounds}.
Under a blind protocol, annotators are not shown the dataset split, generation model, or each
other's decisions.

Each annotator assigns a binary label indicating whether
(1) the question is answerable from the source paper, and
(2) the answer is factually correct and explicitly supported by the paper, without relying on
speculation or external knowledge.
We report pass rates pooled across both annotators and inter-annotator agreement measured by
Cohen's $\kappa$.
The resulting agreement is $\kappa = 0.87$, indicating strong agreement.



\begin{center}
\footnotesize
\setlength{\tabcolsep}{3pt}
\renewcommand{\arraystretch}{1.02}
\resizebox{\columnwidth}{!}{%
\begin{tabular}{lccc}
\toprule
\textbf{Research Component}
& \textbf{Annotator A (Pass / 20)}
& \textbf{Annotator B (Pass / 20)}
& \textbf{Pooled Pass Rate} \\
\midrule
Research Domain            & $17/20$  & $18/20$ & 87.5\% \\
Previous Methods           & $20/20$  & $20/20$ & 100\%  \\
Existing Challenges        & $20/20$  & $20/20$ & 100\%  \\
Motivation                 & $18/20$  & $18/20$ & 90\%   \\
Findings / Assumptions     & $19/20$  & $19/20$ & 95\%   \\
Methods                    & $18/20$  & $18/20$ & 90\%   \\
Experimental Settings      & $19/20$  & $18/20$ & 92.5\% \\
Experimental Results       & $17/20$  & $16/20$ & 82.5\% \\
\midrule
\textbf{Overall}
& $148/160$
& $147/160$
& \textbf{92.2\%} \\
\bottomrule
\end{tabular}}
\captionsetup{hypcap=false}\captionof{table}{Expert-based validation results; each component includes 20 Q\&A pairs sampled
with seed 2026, with pass rates pooled across two annotators.}
\label{tab:expert-validation-two-reviewers}
\end{center}

\section{Human Verification of Judge Reliability}
\label{sec:human-validation}

We assess LLM-as-a-Judge reliability with a blind ranking study on 80 questions, sampled as 10 per
research component.
For each question, a human expert ranks responses from \texttt{o3}, \texttt{DeepSeek-R1}, and
\texttt{Gemini 2.0 Flash} according to factual correctness and evidence grounding.
In parallel, each LLM judge scores the same responses using the rubric above, and we derive
judge-induced rankings from these scores.
Mean Kendall's $\tau$ between human and judge-induced rankings ranges from 0.76 to 0.81
(Table~\ref{tab:judge-human-agreement}).

\begin{center}
\centering
\small
\begin{tabular}{lc}
\toprule
\textbf{Judge Model} & \textbf{Mean Kendall's $\tau$} \\
\midrule
GPT-4.1       & 0.81 \\
DeepSeek-R1   & 0.79 \\
Gemini 2.5    & 0.76 \\
\bottomrule
\end{tabular}
\captionsetup{hypcap=false}\captionof{table}{Rank correlation between human expert rankings and LLM judges.}
\label{tab:judge-human-agreement}
\end{center}

\section{Training Hyperparameters}\label{sec:training-hparams}
\paragraph{SFT} We use \textbf{LLaMA-Factory}\citep{zheng2024llamafactory} for three-epoch SFT and
select the checkpoint with the lowest validation loss. A cosine learning-rate schedule with a 10\%
warmup ratio is applied throughout training; key hyperparameters are summarized in
Table~\ref{tab:hparams-sft}.

\begin{center}
\begin{minipage}{0.72\columnwidth}
  \centering
  \scriptsize
  \renewcommand{\arraystretch}{0.95}
  \setlength{\tabcolsep}{5pt}
  \begin{tabular}{ll}
    \toprule
    \textbf{Hyperparameter} & \textbf{Value} \\
    \midrule
    Learning rate (LR) & $1\times10^{-5}$ \\
    LR scheduler & cosine \\
    Warmup ratio & 0.1 \\
    Batch size (per device) & 4 \\
    Gradient accumulation steps & 8 \\
    Epochs & 3 \\
    Weight decay & 0.01 \\
    \bottomrule
  \end{tabular}
  \captionsetup{hypcap=false}\captionof{table}{Hyperparameter settings for SFT training.}
  \label{tab:hparams-sft}
\end{minipage}
\end{center}

\begin{table}[!ht]
\centering
\scriptsize
\renewcommand{\arraystretch}{0.92}
\setlength{\tabcolsep}{4pt}
\caption{Key hyperparameters for GRPO training using VERL.}
\label{tab:hparams-verl}
\begin{tabular}{@{}ll@{}}
\toprule
\textbf{Hyperparameter} & \textbf{Value} \\
\midrule
\multicolumn{2}{@{}l}{\textit{Data Configuration}} \\
Max prompt length & 1024 \\
Max response length & 4096 \\
\midrule
\multicolumn{2}{@{}l}{\textit{Algorithm Configuration}} \\
Advantage estimator & \texttt{grpo} \\
KL coefficient ($\beta_{\text{KL}}$) & $1.0\times10^{-3}$ \\
Seed & 1 \\
\midrule
\multicolumn{2}{@{}l}{\textit{Worker: Actor Configuration}} \\
Global batch size & 1024 \\
Micro-batch (update, per device) & 8 \\
Mini-batch (per update) & 16 \\
Learning rate (actor) & $1.0\times10^{-6}$ \\
Optimizer & Adam \\
LR warmup ratio & 0.0 \\
\midrule
\multicolumn{2}{@{}l}{\textit{Worker: Rollout Configuration}} \\
Sampler backend & vLLM \\
Rollout trajectories ($n$) & 6 \\
Temperature & 1.0 \\
Top-$p$ & 0.7 \\
Tensor model parallel size & 2 \\
\midrule
\multicolumn{2}{@{}l}{\textit{Trainer Configuration}} \\
Total epochs & 3 \\
Nodes & 1 \\
GPUs per node & 8 \\
\bottomrule
\end{tabular}
\end{table}

\paragraph{GRPO} We use \textbf{VERL}\citep{sheng2024hybridflow} for three-epoch GRPO, initializing
the actor from the SFT checkpoint and generating vLLM rollouts with $n=6$ responses per prompt.
KL regularization uses $\beta_{\text{KL}} = 10^{-3}$ on a single node with 8 GPUs.

\section{Additional Dataset Statistics}\label{appendix:data-stats}

Table~\ref{tab:conference-stats} summarizes the source pool across six AI, ML, and CV conferences.
After Creative Commons license filtering, 14,474 papers are retained for dataset construction.

\begin{center}
\begin{minipage}{0.58\columnwidth}
\centering
\captionsetup{hypcap=false}\captionof{table}{Statistics of dataset sources.}
\label{tab:conference-stats}

\vspace{0.2em}
\tablestyle{5pt}{0.92}
\begin{tabular}{l r}
\toprule
Conference & Count \\
\midrule
NeurIPS & 20,286 \\
ICML    & 10,979 \\
ICLR    & 11,679 \\
CVPR    & 11,842 \\
ICCV    & 5,369  \\
ECCV    & 6,166  \\
\bottomrule
\end{tabular}
\end{minipage}
\end{center}

\begin{center}
\begin{minipage}{\columnwidth}
\centering

\begin{minipage}[t]{0.47\linewidth}
  \centering
  \includegraphics[width=0.97\linewidth]{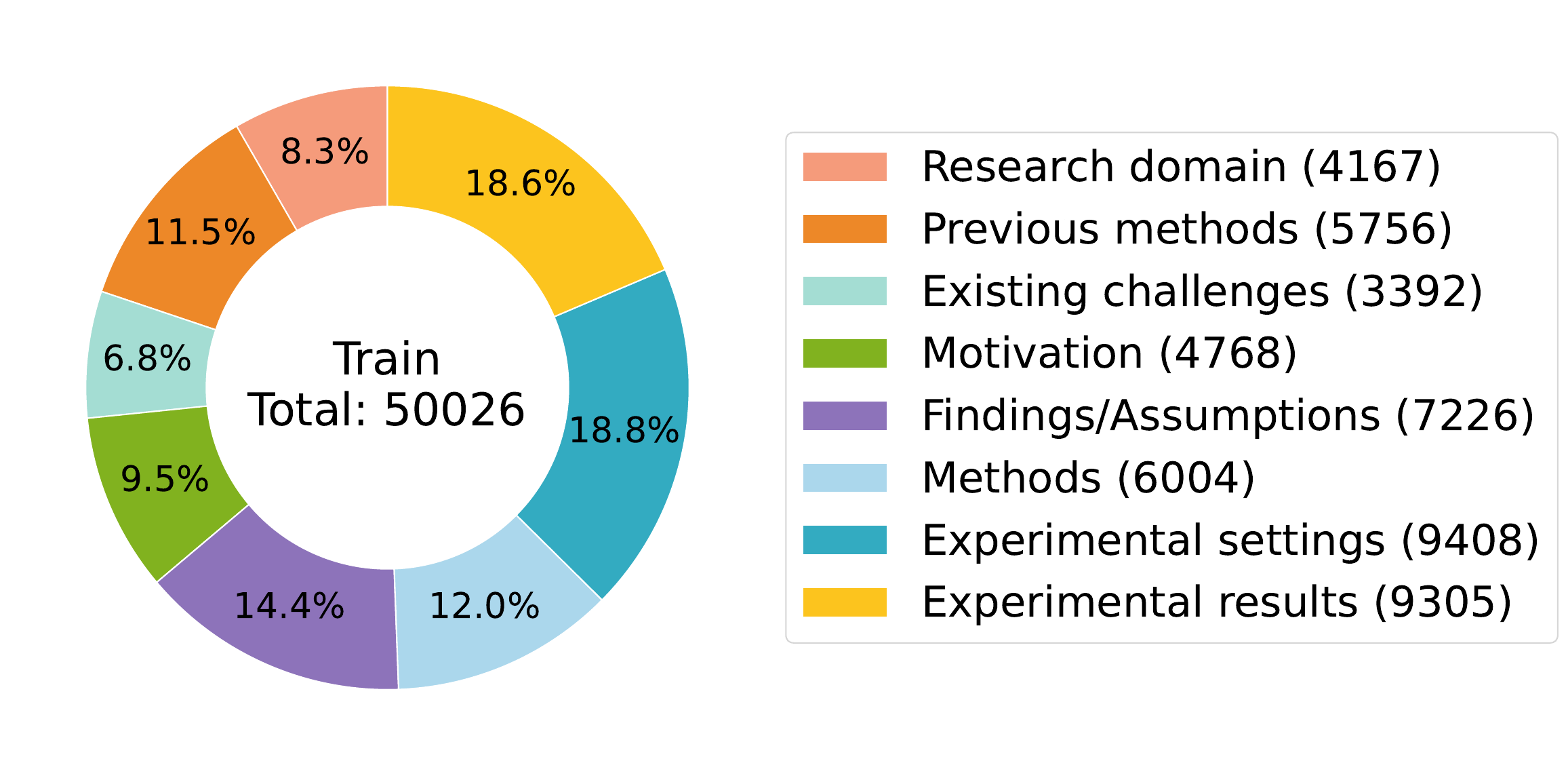}
  \parbox{\linewidth}{\centering\footnotesize (a) Category distribution (CS-50k / train)}
\end{minipage}\hfill
\begin{minipage}[t]{0.47\linewidth}
  \centering
  \includegraphics[width=0.97\linewidth]{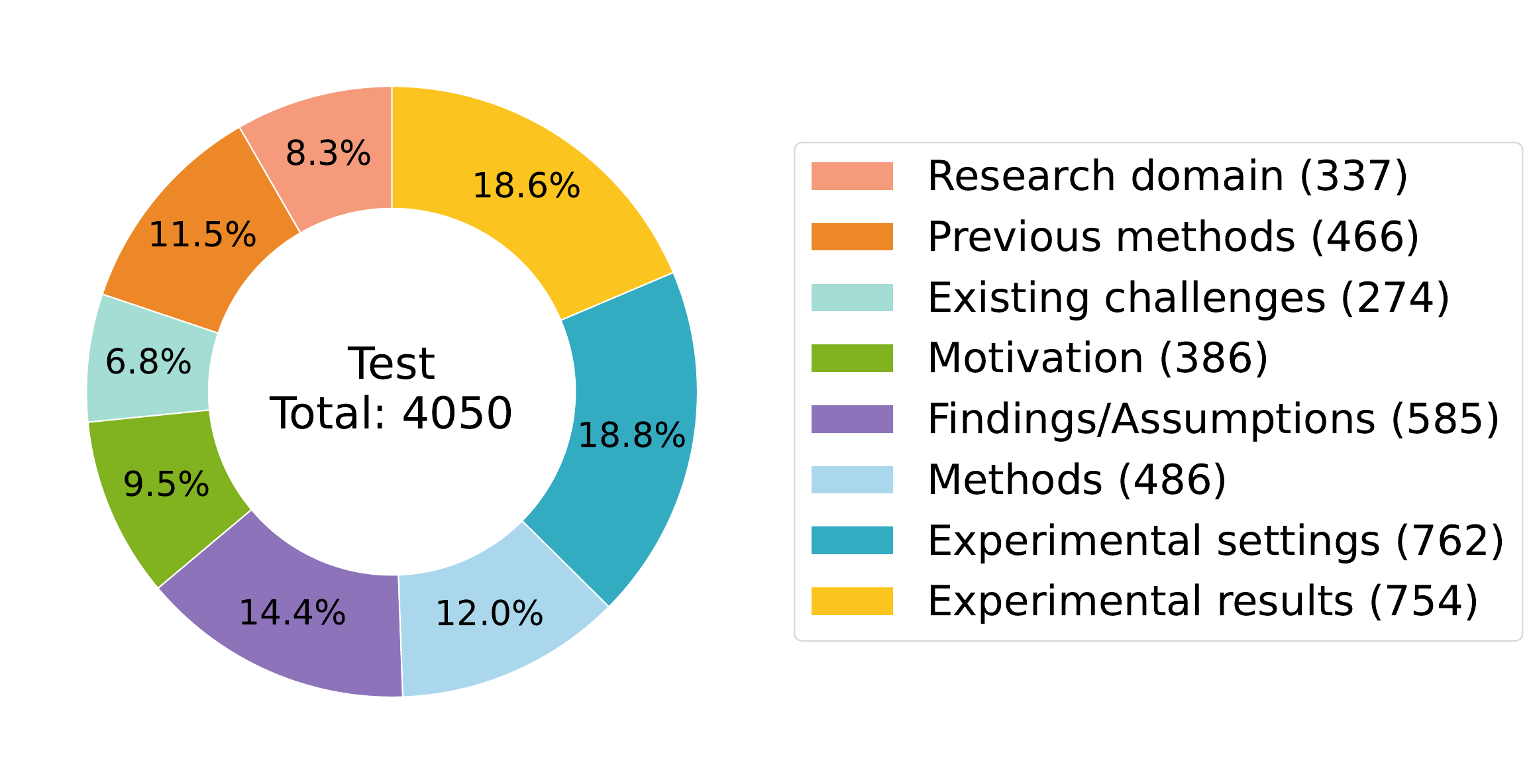}
  \parbox{\linewidth}{\centering\footnotesize (b) Category distribution (CS-4k / evaluation)}
\end{minipage}

\vspace{0.45em}

\begin{minipage}[t]{0.47\linewidth}
  \centering
  \includegraphics[width=0.97\linewidth]{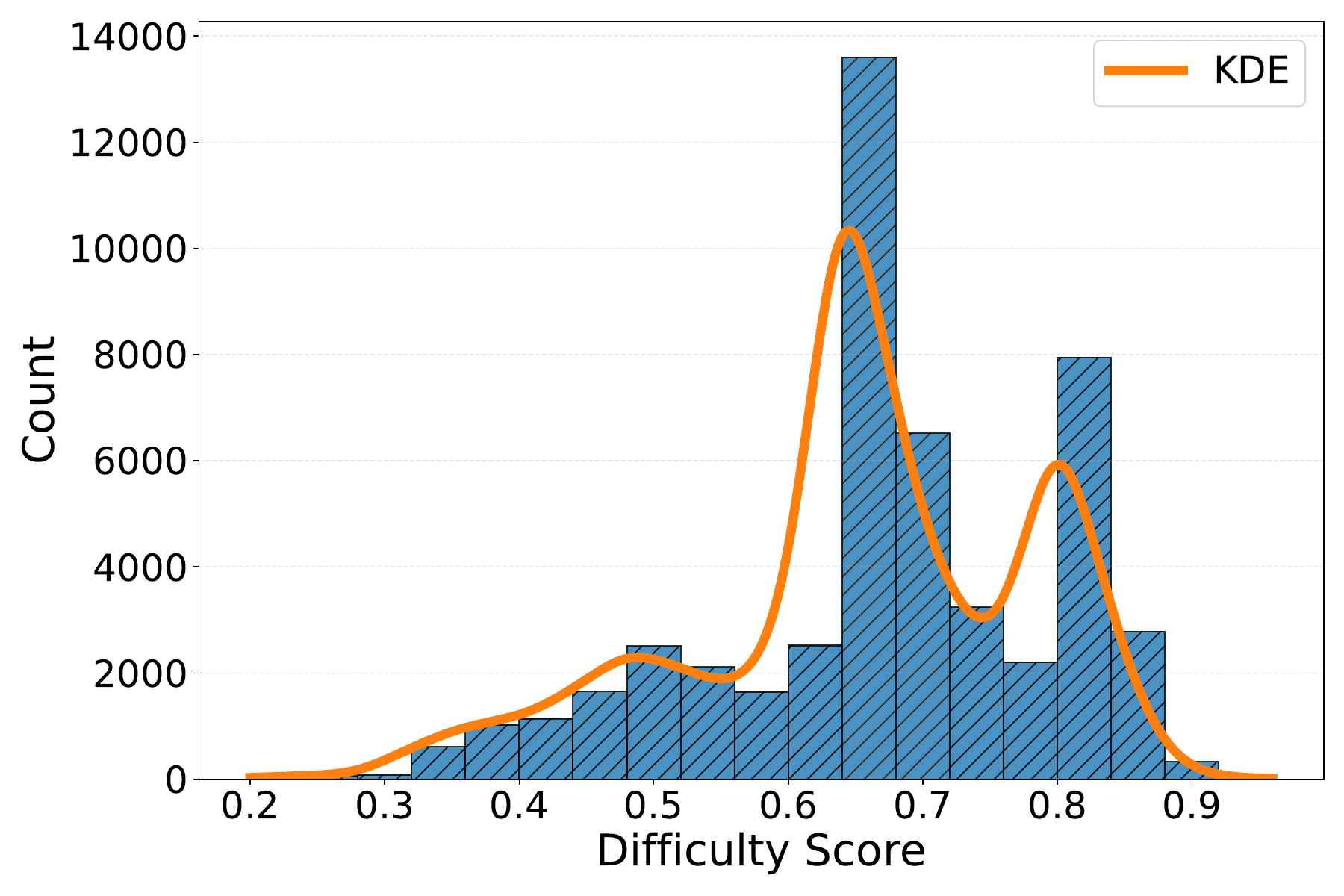}
  \parbox{\linewidth}{\centering\footnotesize (c) Difficulty (CS-50k / train)}
\end{minipage}\hfill
\begin{minipage}[t]{0.47\linewidth}
  \centering
  \includegraphics[width=0.97\linewidth]{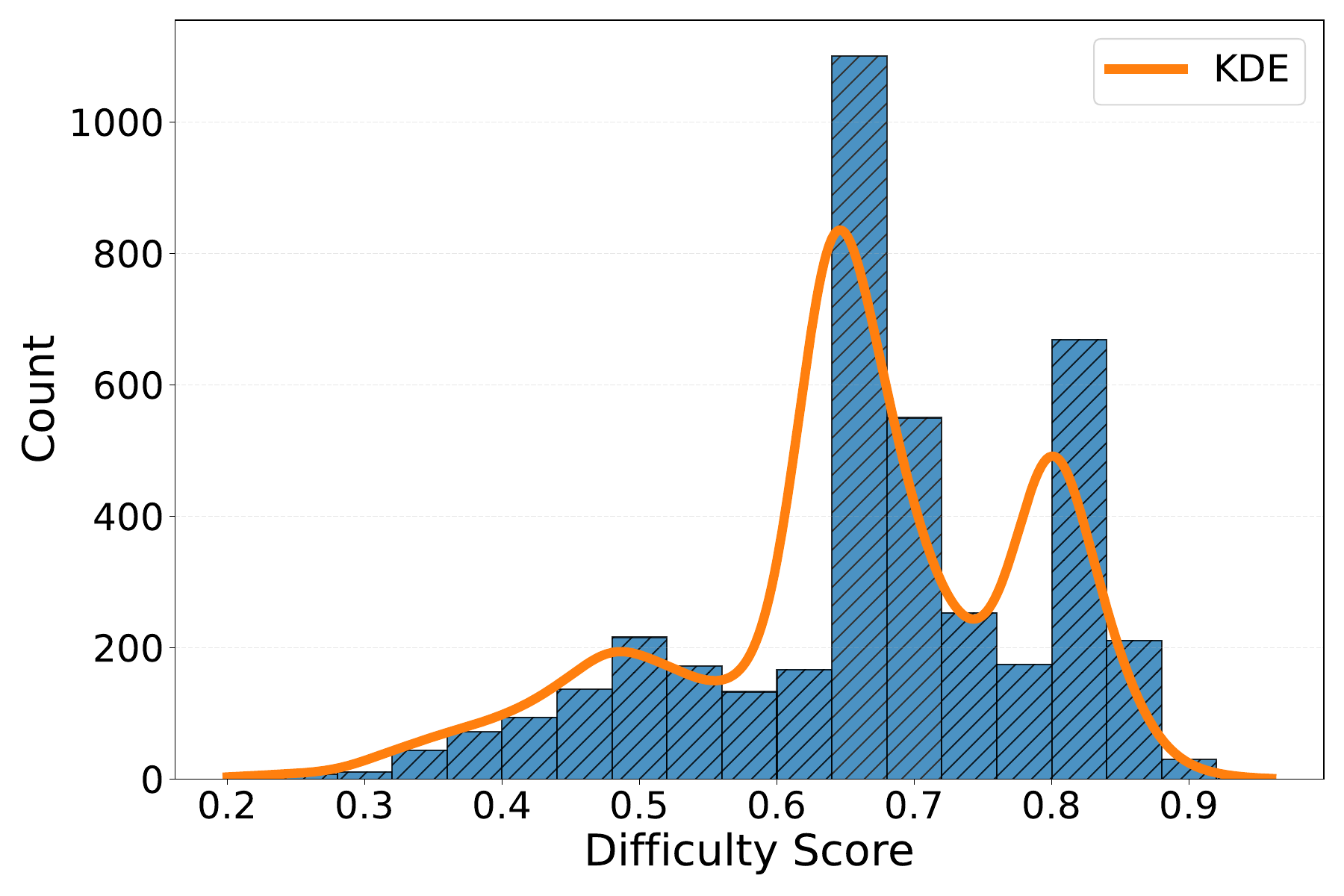}
  \parbox{\linewidth}{\centering\footnotesize (d) Difficulty (CS-4k / evaluation)}
\end{minipage}

\vspace{0.45em}

\begin{minipage}[t]{0.47\linewidth}
  \centering
  \includegraphics[width=0.97\linewidth]{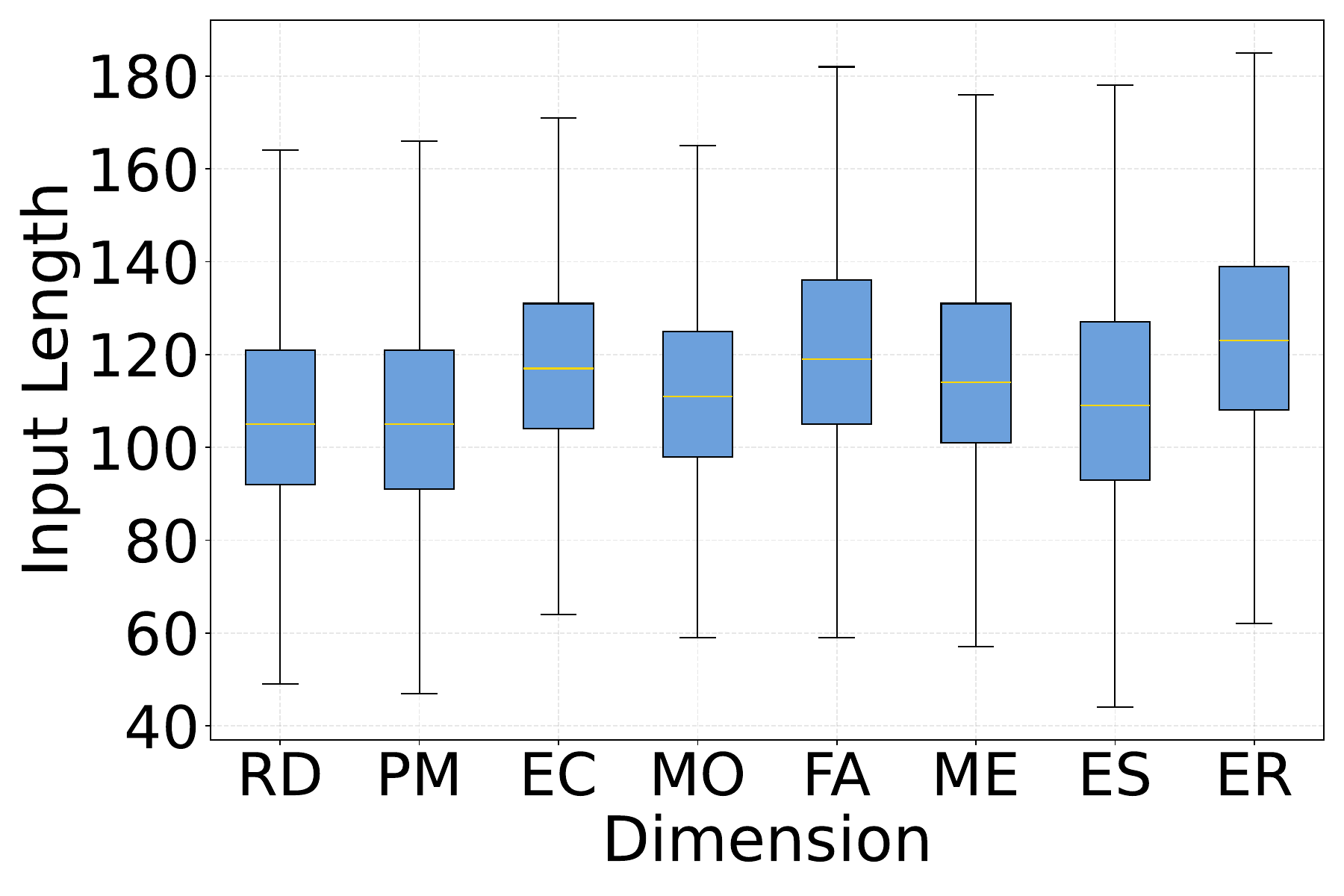}
  \parbox{\linewidth}{\centering\footnotesize (e) Input length (CS-50k / train)}
\end{minipage}\hfill
\begin{minipage}[t]{0.47\linewidth}
  \centering
  \includegraphics[width=0.97\linewidth]{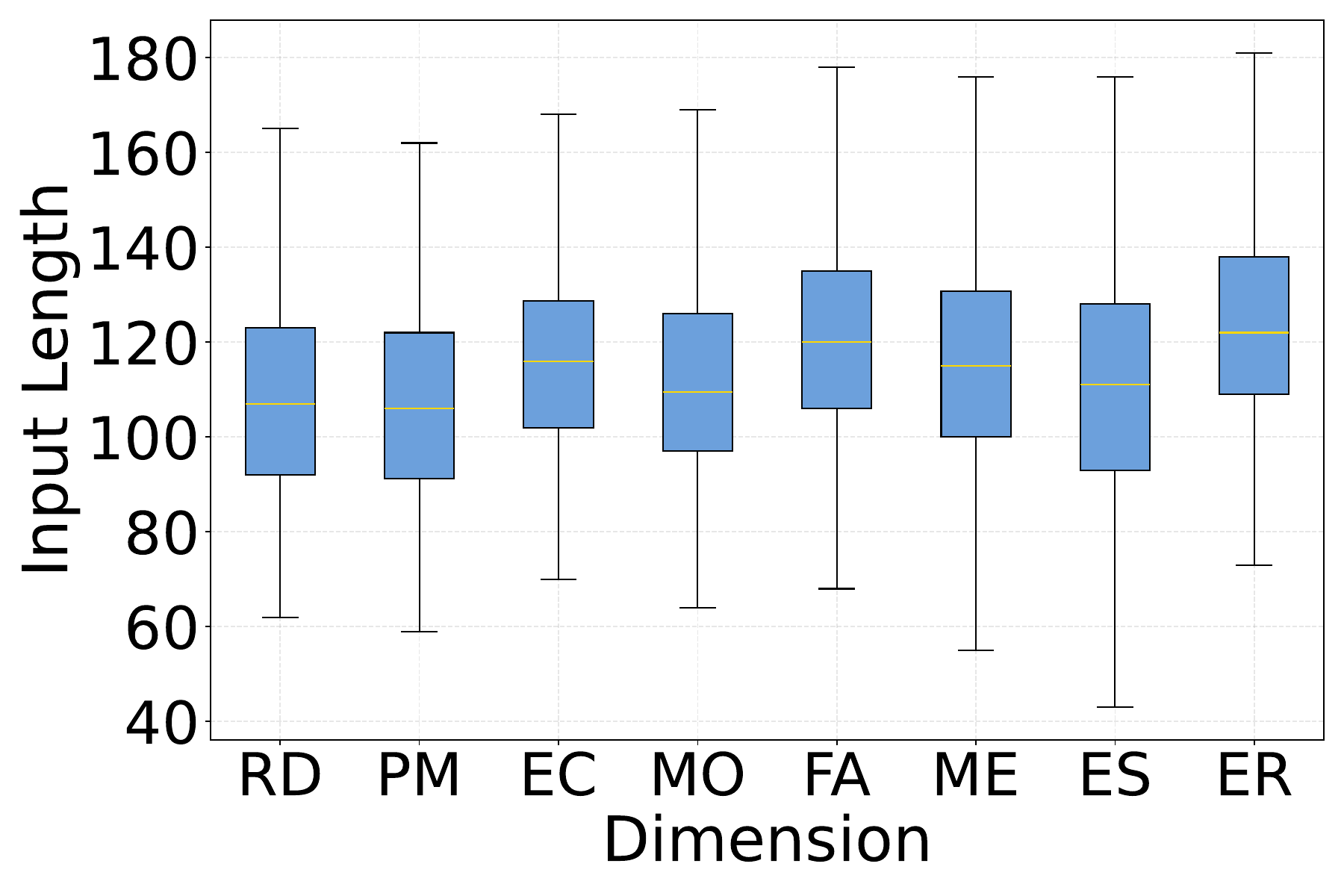}
  \parbox{\linewidth}{\centering\footnotesize (f) Input length (CS-4k / evaluation)}
\end{minipage}

\captionsetup{hypcap=false}\captionof{figure}{Overview of category, difficulty, and input-length distributions for train (CS-50k) and
evaluation (CS-4k).}
\label{fig:all-stats}
\end{minipage}
\end{center}

We report category, difficulty, and input length distributions for CS-50k (training) and CS-4k
(question-level evaluation), showing comparable coverage with the evaluation subset slightly skewed
toward harder examples
(Figures~\ref{fig:all-stats}).

\end{document}